%% file: main.tex
\pdfoutput=1
\documentclass[10pt,twocolumn,letterpaper]{article}

\usepackage[pagenumbers]{cvpr}

\input{preamble}

\definecolor{cvprblue}{rgb}{0.21,0.49,0.74}
\usepackage[pagebackref,breaklinks,colorlinks,allcolors=cvprblue]{hyperref}
\usepackage{caption}
\usepackage[capitalize]{cleveref}
\hypersetup{pdftitle={TacInno: Teaching Vision-Language-Action Models the Contact Dynamics Beyond Vision and Action},
  pdfauthor={Weile Li, Yunfan Lou, Muqing Shi, Zihao Zhang, Zhenbiao Huang, Boxian Yao, Xiyu Zhu, Zhibo Pang}}
\crefname{section}{Sec.}{Secs.}
\Crefname{section}{Section}{Sections}
\crefname{table}{Tab.}{Tabs.}
\Crefname{table}{Table}{Tables}
\crefname{figure}{Fig.}{Figs.}
\Crefname{figure}{Figure}{Figures}
\crefname{equation}{Eq.}{Eqs.}
\Crefname{equation}{Equation}{Equations}

\title{\method: Teaching Vision-Language-Action Models\\the Contact Dynamics Beyond Vision and Action}

\author{%
Weile Li\textsuperscript{1,2*}\quad
Yunfan Lou\textsuperscript{3*}\quad
Muqing Shi\textsuperscript{4*}\quad
Zihao Zhang\textsuperscript{2}\\
Zhenbiao Huang\textsuperscript{5}\quad
Boxian Yao\textsuperscript{3}\quad
Xiyu Zhu\textsuperscript{3}\quad
Zhibo Pang\textsuperscript{3\dag}\\[0.5em]
{\normalsize\textsuperscript{1}ShanghaiTech University\quad
\textsuperscript{2}Institute of AI for Industries (IAII), Chinese Academy of Sciences}\\
{\normalsize\textsuperscript{3}Peking University\quad
\textsuperscript{4}Washington University in St. Louis\quad
\textsuperscript{5}Beijing FiveAges Technology}\\[0.3em]
{\tt\small zhibo.pang@pku.edu.cn}\\[0.5em]
{\normalsize Project page: \href{https://tacinno.github.io}{\texttt{tacinno.github.io}}}
}

\begin{document}
\twocolumn[{%
\renewcommand\twocolumn[1][]{#1}%
\maketitle
\begin{center}
  \centering
  \captionsetup{type=figure}
  \includegraphics[width=\textwidth]{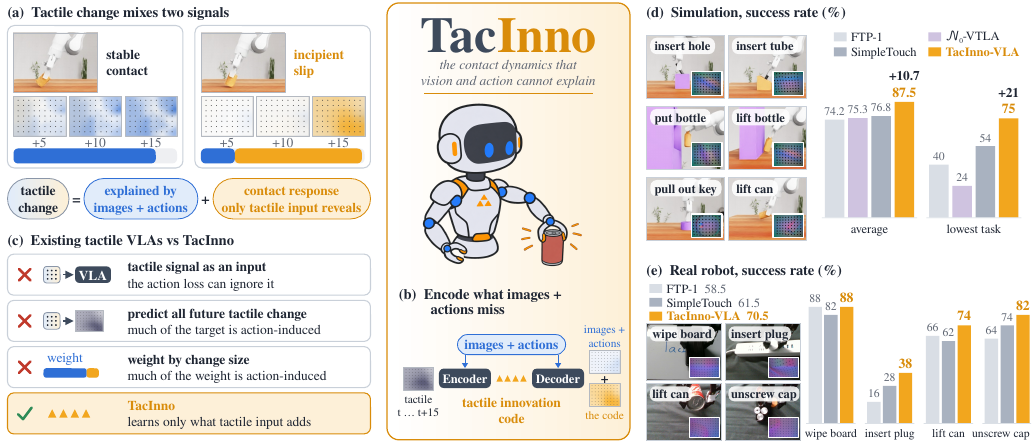}
  \captionof{figure}{\textbf{Overview of \method.} (a) The tactile change mixes a component explained by vision and executed actions with a contact response that only tactile input reveals. (b) The \method tokenizer encodes the contact response into the tactile innovation code. (c) Existing visuo-tactile VLA models versus \method, which uses the code solely as a training target. (d) Average and lowest per-task success rates on six UniVTAC tasks. (e) Success rates on four real-world tasks.}
  \label{fig:teaser}
\end{center}
\vspace{1em}
}]
{\let\thefootnote\relax\footnotetext{* Equal contribution. \quad \dag\ Corresponding author.}}
\input{sec/0_abstract}
\input{sec/1_intro}
\input{sec/2_related}
\input{sec/3_method}
\input{sec/4_experiments}
\input{sec/5_conclusion}
\input{sec/6_llm_statement}
{
    \small
    \bibliographystyle{ieeenat_fullname}
    \bibliography{main}
}

\clearpage
\renewcommand{\dbltopfraction}{0.95}
\renewcommand{\dblfloatpagefraction}{0.8}
\setcounter{dbltopnumber}{2}
\maketitlesupplementary
\appendix
\input{sec/S_supp}

\end{document}

%% file: preamble.tex
\usepackage{graphicx}
\graphicspath{{figs/}}
\usepackage{amsmath,amssymb}
\usepackage{booktabs}
\usepackage{multirow}
\usepackage{xcolor}
\usepackage{xspace}
\newcommand{\method}{TacInno\xspace}
\newcommand{\vla}{TacInno-VLA\xspace}

\newcommand{\ih}{Insert Hole\xspace}
\newcommand{\itb}{Insert Tube\xspace}
\newcommand{\lb}{Lift Bottle\xspace}
\newcommand{\lc}{Lift Can\xspace}
\newcommand{\pok}{Pull-out Key\xspace}
\newcommand{\pb}{Put Bottle\xspace}
\newcommand{\nzero}{$\mathcal{N}_0$-VTLA\xspace}
\newcommand{\dX}{\Delta X}
\newcommand{\E}{\mathbb{E}}

%% file: sec/0_abstract.tex
\begin{abstract}
Vision-Language-Action (VLA) models have recently been equipped with tactile sensing to perceive visually subtle contact events in contact-rich manipulation. However, adding tactile input yields inconsistent gains, and some visuo-tactile VLA models rely little on tactile input, because much of the tactile change, such as the gel deformation during gripper closure, is predictable from visual observations and executed actions. To address this problem, we introduce \method, which isolates the contact dynamics that vision and executed actions cannot explain, termed \emph{tactile innovation}, and uses tactile innovation only as a training target. Specifically, the \method tokenizer, conditioned on visual observations and executed actions, compresses the future tactile change into a discrete code, and the policy, \vla, learns to predict the code, adding no module for the code at deployment. \vla achieves average success rates of 87.5\% across six simulation tasks and 70.5\% across four real-world tasks, outperforming the strongest visuo-tactile baseline by 10.7 and 9.0 percentage points, respectively. Ablation studies validate the contributions of visual and action conditioning, multi-frame encoding, and auxiliary code prediction. Together, these results demonstrate the effectiveness of tactile innovation as a supervision signal for contact-rich manipulation.
\end{abstract}

%% file: sec/1_intro.tex
\section{Introduction}
\label{sec:intro}

Vision-Language-Action (VLA) models have emerged as a promising approach to general-purpose robotic manipulation~\cite{brohan2023rt2,kim2024openvla,black2024pi0,bjorck2025gr00t}. Building on pretrained Vision-Language Models (VLMs), VLA models learn to predict robot actions from visual observations, language instructions, and proprioceptive states through imitation learning~\cite{intelligence2025pi05}. Nevertheless, contact-rich manipulation remains challenging, as critical contact events, such as incipient slip and jamming during insertion, are often difficult to infer from external visual observations. To capture contact events, recent efforts augment VLA models with tactile observations~\cite{yuan2026ftp1,huang2025tactilevla,simpletouch2026,zhang2026restacvla,neoteai2026n0vtla,zhang2025vtla,cheng2026omnivtla,bi2026vlatouch,li2026atvla}, typically acquired by vision-based tactile sensors such as GelSight~\cite{yuan2017gelsight} and DIGIT~\cite{lambeta2020digit}, which capture the deformation of a soft gel.

Existing visuo-tactile policies incorporate tactile observations as additional inputs~\cite{huang2025tactilevla,tapvla2026,yuan2026ftp1,tacvla2026,tacforcing2026}, predict future tactile signals~\cite{chen2026ratg,fu2026decal,wang2026tacdynwam,simpletouch2026,zhang2026contactworld}, or emphasize tactile changes during training~\cite{tau2026vla,dreamtac2026,tactilewam2026,hitacwam2026,medex2026,Xu2026VTMUSEMU} (\cref{fig:teaser}c). However, incorporating tactile inputs yields inconsistent gains and can even reduce the success rate relative to the same VLA without tactile inputs~\cite{yuan2026ftp1,zhang2026restacvla}. Moreover, we apply a tactile-off test, which removes tactile input at test time, and find that the success rates of some existing policies change little, indicating limited reliance on tactile information (\cref{sec:touchoff}). We attribute the inconsistent gains and the limited reliance to the nature of the tactile signal, since a tactile sensor measures how the contact responds to the robot's own actions. Consequently, given the initial tactile observation, a substantial portion of the subsequent tactile change, such as the gel deformation during gripper closure, is predictable from visual observations and executed actions (\cref{sec:attribution}, \cref{fig:teaser}a). As an input, the predictable portion is largely redundant, and adding an input modality can even degrade performance~\cite{wang2020multimodal}. When future tactile signals are the prediction target, the predictable portion makes up most of the target, so the prediction loss is dominated by changes that vision and executed actions already explain. Weighting the loss by the magnitude of tactile change does not help, since the largest changes are often the predictable ones (\cref{fig:timeline}). A central question arises: \emph{How can the contact dynamics that vision and executed actions cannot explain be isolated and used to guide VLA policy learning?}

To answer this question, we propose \method, which isolates the contact dynamics that vision and executed actions cannot explain as the supervision signal for VLA training. Inspired by the notion of innovation in estimation theory~\cite{kalman1960filtering,kailath1968innovations}, we define \emph{tactile innovation} as the residual tactile change after accounting for the initial tactile observation, visual observations, and executed actions. Three properties of tactile innovation determine the design of \method. First, tactile innovation is \emph{conditional} on visual observations and executed actions. We therefore introduce the TacInno tokenizer, which learns a compact discrete code as a proxy for tactile innovation and conditions both its encoder and decoder on the corresponding visual observations and executed actions, so that the code retains the contact dynamics that the two conditions cannot explain (\cref{fig:teaser}b). Second, tactile innovation is \emph{dynamic}, arising from transient contact events, such as incipient slip, that span several control steps. The tokenizer therefore encodes a short sequence of tactile observations with causal temporal attention rather than a single pair of frames. Third, tactile innovation lies in the \emph{future} and is unavailable when the policy acts. We therefore train \vla to predict the code from current observations as an auxiliary objective, supervising the tactile representations used for action prediction. The code serves only as a training target, and the tokenizer and auxiliary prediction heads are removed at deployment.

We evaluate \method on six contact-rich tasks from the UniVTAC benchmark~\cite{univtac2026} and four real-world tasks using a Franka FR3 robot equipped with a vision-based tactile sensor. \vla achieves average success rates of 87.5\% in simulation and 70.5\% in the real world, the highest among the evaluated visuo-tactile policies. Ablation studies verify the contribution of each design, and the tactile-off test confirms that \vla relies on tactile input.

Our contributions are summarized as follows:
\begin{itemize}
    \item With the tactile-off test, we show that some existing visuo-tactile policies rely little on tactile input, attribute the limited reliance to the redundancy of the tactile signal, and propose to separate the tactile change that vision and executed actions cannot explain, termed tactile innovation.
    \item We propose \method, which encodes tactile innovation into a discrete code with a conditional multi-frame tokenizer and uses the code solely as an auxiliary training target, adding no module for the code at deployment.
    \item \vla outperforms the strongest visuo-tactile baseline by 10.7 and 9.0 percentage points in average success rate in simulation and the real world, respectively. The success rate of \vla drops from 88\% to 0\% on \lc under the tactile-off test and by 21 percentage points on the lifting tasks when the code encodes the entire tactile change.
\end{itemize}

%% file: sec/2_related.tex
\section{Related Work}
\label{sec:related}

\paragraph{Visuo-tactile policy learning.}
Earlier visuo-tactile policies fuse tactile images~\cite{lee2019making,huang20243dvitac,liu2025vitamin} or pretrained tactile representations~\cite{higuera2024sparsh,zhao2024t3,Xu2026VTMUSEMU} with visual features in imitation-learning policies such as ACT~\cite{zhao2023act} and Diffusion Policy~\cite{chi2023diffusion}. Recent methods extend tactile sensing to VLA models in three progressive directions (\cref{fig:teaser}c). The first incorporates tactile observations as additional inputs and relies on the action loss alone to determine how the tactile inputs are used~\cite{yuan2026ftp1,huang2025tactilevla,tacvla2026,tapvla2026,tacforcing2026,neoteai2026n0vtla,zhang2025vtla,cheng2026omnivtla,bi2026vlatouch,li2026atvla}. Because the action loss rewards only what improves the imitated actions, the policy is not required to use the tactile inputs, and the gains are modest and occasionally negative~\cite{yuan2026ftp1,zhang2026restacvla}. FACT~\cite{fact2026vla} further replaces the tactile input with noise to examine how much of the reported gain originates from tactile information. The second direction therefore assigns the tactile signal an objective of its own and predicts future tactile signals, either within the policy~\cite{simpletouch2026,chen2026ratg,fu2026decal,liu2026physword} or through tactile world models~\cite{wang2026tacdynwam,zhang2026contactworld}. Since predicting every future tactile signal spends much of the supervision on periods in which the contact hardly changes, the third direction emphasizes tactile changes during training by weighting or gating the tactile signal according to the magnitude of its change~\cite{tau2026vla,medex2026,dreamtac2026,tactilewam2026,hitacwam2026}. However, the second and third directions still supervise the tactile representation with the entire future tactile change, including the component that vision and executed actions already explain. Our method extends this progression by supervising only the component that vision and executed actions cannot explain. The closest work, ResTacVLA~\cite{zhang2026restacvla}, uses the quantized residual between visually predicted and observed tactile representations as a policy input, whereas our code encodes the future tactile change that vision and executed actions cannot explain and serves only as a training target.

\paragraph{Discrete latent dynamics learning.}
Vector-quantized variational autoencoders (VQ-VAEs)~\cite{oord2017vqvae} learn compact discrete representations by mapping continuous features to entries in a learned codebook and reconstructing observations from the quantized codes. Vector quantization has been extended to temporal modeling, where discrete latent variables capture transitions between observations. Genie~\cite{bruce2024genie} learns latent actions from unlabeled videos to enable controllable video generation. LAPA~\cite{ye2025lapa} employs a VQ-VAE-based action tokenizer to extract discrete latent actions from video transitions, which serve as supervision for VLA pretraining. UniVLA~\cite{bu2025univla} further incorporates language conditioning into inverse- and forward-dynamics learning to obtain task-centric latent actions. Building on discrete latent dynamics learning, our method applies vector quantization to tactile observations. The encoder compresses a sequence of tactile observations into discrete codes, while the decoder reconstructs future tactile changes conditioned on the current tactile state, visual observations, and executed actions. Unlike latent action models, which infer unobserved actions, our tokenizer is trained on demonstrations with known actions and therefore provides the actions to both the encoder and the decoder. With conditional reconstruction, the codes retain only the contact dynamics that the conditioning information cannot explain.

%% file: sec/3_method.tex
\section{Method}
\label{sec:method}

\begin{figure*}[t]
  \centering
  \includegraphics[width=\textwidth]{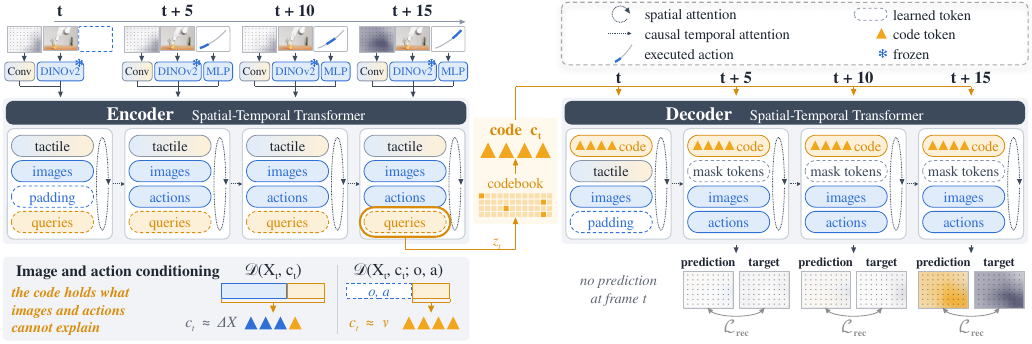}
  \caption{\textbf{Architecture of the \method tokenizer.} The encoder compresses the tactile observations from $t$ to $t{+}15$ into the 16-bit code $c_t$, and the decoder reconstructs the future tactile change from $X_t$ and $c_t$. The encoder and the decoder are both conditioned on visual observations and executed actions, so that $c_t$ retains tactile innovation (lower left).}
  \label{fig:pipeline}
\end{figure*}

In this section, we present \method in three parts (\cref{fig:pipeline,fig:pi05}). First, \cref{sec:formulation} formulates the problem and defines tactile innovation. Then, \cref{sec:teacher} presents the \method tokenizer, which isolates tactile innovation from demonstrations and labels each demonstration frame with a discrete code. Finally, \cref{sec:student} describes \vla, which is trained to predict the code as an auxiliary objective.

\subsection{Problem Formulation}
\label{sec:formulation}
At control step $t$, the policy receives visual observations $o_t$, a language instruction $\ell$, the proprioceptive state $q_t$, and the tactile observation $X_t$, which comprises the indentation depth map and the marker displacement field of the vision-based tactile sensors. The policy predicts an action chunk $a_{t+1:t+H}$ over the next $H$ steps, where $a_{t+i}$ denotes the offset of the joint target at step $t{+}i$ from $q_t$. The policy is trained by imitation learning on expert demonstrations.

\paragraph{Tactile innovation.}
To characterize the tactile change after step $t$, we sample four time steps, $t$, $t{+}5$, $t{+}10$, and $t{+}15$, and index the last three as $t{+}5k$ with $k\in\{1,2,3\}$. Visual and tactile observations are sampled only at the four time steps, denoted $o_{t:t+5k}=(o_t,o_{t+5},\dots,o_{t+5k})$, whereas actions are taken at every control step, denoted $a_{t+1:t+5k}=(a_{t+1},\dots,a_{t+5k})$. The cumulative tactile change is defined as
\begin{equation}
  \dX_{t:t+5k} \;=\; X_{t+5k}-X_t,\qquad k\in\{1,2,3\}.
  \label{eq:dx}
\end{equation}
Conditioned on the current tactile observation $X_t$, the visual observations $o_{t:t+5k}$, and the executed actions $a_{t+1:t+5k}$, the cumulative tactile change decomposes as
\predisplaypenalty=0%
\begin{align}
  \dX_{t:t+5k} &= \underbrace{\E\big[\dX_{t:t+5k}\mid X_t,o_{t:t+5k},a_{t+1:t+5k}\big]}_{\text{predictable from vision and actions}} \nonumber\\
  &\quad + \underbrace{\nu_{t:t+5k}}_{\text{tactile innovation}} ,
  \label{eq:innovation}
\end{align}
\predisplaypenalty=10000
where the first term denotes the component predictable from the visual observations and executed actions, and the residual $\nu_{t:t+5k}$ denotes the component that the two conditions cannot explain, which we term tactile innovation. Since the conditions include the visual observations up to $t{+}5k$, a contact response that has become visible is explained by the first term, and $\nu$ retains only the contact responses that the visual observations have not yet revealed. Future visual observations are included because a closed-loop policy perceives any contact response that becomes visible within the window through its cameras at subsequent steps, so tactile sensing only needs to provide what the visual observations have not yet revealed.

\subsection{TacInno Tokenizer}
\label{sec:teacher}
Tactile innovation cannot be computed directly from demonstrations, because the conditional expectation in \cref{eq:innovation} is unknown and must be estimated from a finite set of demonstrations. Moreover, even if $\nu$ were available, $\nu$ would be a per-pixel change field over both tactile sensors with the same dimensionality as $X_t$, which is too high-dimensional to serve as a prediction target for the policy. We therefore learn a 16-bit discrete code $c_t=(c_t^{(1)},\dots,c_t^{(4)})\in\{0,\dots,15\}^4$ as a compact proxy for $\nu$, comprising four tokens that each take one of 16 values.

\paragraph{Tokenizer architecture.}
\Cref{fig:pipeline} illustrates the overall architecture of the tokenizer. Since isolating $\nu$ is equivalent to compressing the information that a window of tactile observations carries beyond the visual observations and executed actions, we adopt the encoder-decoder structure of latent action models~\cite{bruce2024genie,bu2025univla,ye2025lapa}, which can be compactly expressed as
\begin{equation}
  \begin{aligned}
    z_t &= \mathcal{E}\big(X_{t:t+15};\,o_{t:t+15},a_{t+1:t+15}\big),\\
    c_t &= \mathrm{VQ}(z_t),\\
    \{\widehat{\dX}_{t:t+5k}\}_{k=1}^{3} &= \mathcal{D}\big(X_t,c_t;\,o_{t:t+15},a_{t+1:t+15}\big),
  \end{aligned}
  \label{eq:encdec}
\end{equation}
where the encoder $\mathcal{E}$, an inverse dynamics model, takes the four tactile observations of the window as input and compresses the tactile change into four latent vectors $z_t=(z_t^{(1)},\dots,z_t^{(4)})$. $\mathrm{VQ}$ denotes vector quantization~\cite{oord2017vqvae} with a separate codebook $\mathcal{C}_s$ of 16 learnable vectors for each token $s$, and $c_t^{(s)}$ is the index of the vector in $\mathcal{C}_s$ nearest to $z_t^{(s)}$. The codebooks are trained jointly with the encoder and the decoder under the VQ-VAE objective. The decoder $\mathcal{D}$, a forward dynamics model, then reconstructs the tactile changes at the three future time steps from the current tactile observation $X_t$ and the code $c_t$ in a single forward pass. Since the inputs after the semicolons serve as conditions and temporal attention is causal, the $k$-th output is conditioned only on $o_{t:t+5k}$ and $a_{t+1:t+5k}$, as in \cref{eq:innovation}.

\paragraph{Visual and action conditioning.}
The conditions determine the content of the code. Information that the decoder receives directly need not be encoded in the code. Given $X_t$ alone, the code would have to encode the entire tactile change, whereas a decoder that also receives the visual observations and executed actions has access to all variables on which the first term of \cref{eq:innovation} depends. Since the decoder receives no tactile observation after $t$, information about $\nu$ reaches the decoder only through $c_t$, and the reconstruction objective thus encourages the 16-bit code to encode $\nu$~\cite{alemi2017vib}. The encoder receives the same conditions, which allows the encoder to identify the tactile changes that the conditions already explain.

\paragraph{Implementation and training.}
We implement $\mathcal{E}$ and $\mathcal{D}$ as spatial-temporal Transformers~\cite{bruce2024genie} with spatial attention within each frame and causal temporal attention across frames. Each of the four time steps corresponds to one frame composed of tactile, image, and action tokens, where the image features are extracted by a frozen DINOv2~\cite{oquab2024dinov2} (\cref{fig:pipeline}). The encoder appends four learnable queries to each frame. Since temporal attention is causal, only the queries of frame $t{+}15$ attend to the entire window, and the outputs of the queries of frame $t{+}15$ form the four vectors of $z_t$. The decoder reconstructs $\widehat{\dX}_{t:t+5k}$ from learnable mask tokens in the future frames and is supervised by $\dX_{t:t+5k}$ with a contact-weighted $L_1$ loss. The trained tokenizer labels every demonstration frame with a code $c_t$.

\subsection{TacInno-VLA Policy}
\label{sec:student}
\begin{figure}[t]
  \centering
  \includegraphics[width=\columnwidth]{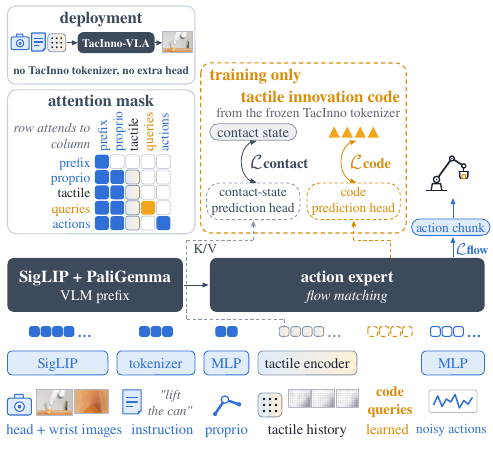}
  \caption{\textbf{Architecture of \vla.} Tactile tokens and four learnable code queries are inserted into the action expert of $\pi_{0.5}$. During training, two auxiliary heads predict the tactile innovation code from the code queries and the current contact state from the tactile tokens. The attention mask (left) prevents all other tokens from attending to the code queries. At deployment (top left), the \method tokenizer and both auxiliary heads are removed.}
  \label{fig:pi05}
\end{figure}

The code $c_t$ of each demonstration frame is provided to the policy as a training target rather than an input, since $c_t$ summarizes future observations that are unavailable when the policy acts. The policy must therefore learn to predict $c_t$ from the inputs available at step $t$, which is feasible because early signs of tactile innovation, such as the shear that increases before slip, often appear in the preceding tactile observations. To this end, we equip the VLA with a tactile branch that encodes three tactile observations and is supervised by the code prediction loss (\cref{fig:pi05}).

\paragraph{Policy architecture.}
\vla is built upon the pretrained VLA $\pi_{0.5}$~\cite{intelligence2025pi05}, which combines a PaliGemma VLM backbone with a Gemma 300M action expert trained with flow matching. The backbone encodes the visual observations and the language instruction into prefix tokens, and the action expert attends to the prefix tokens to generate actions. We insert three groups of condition tokens before the noisy action tokens of the action expert. The three groups are proprioceptive tokens encoded from $q_t$, 60 tactile tokens encoded by a lightweight Transformer tactile encoder from the three tactile observations ($t{-}8$, $t{-}4$, and $t$), and four learnable code queries $u_1,\dots,u_4$, which are used solely to predict the tactile innovation code.

We adopt a block-wise attention mask under which the code queries attend to the prefix, proprioceptive, and tactile tokens, whereas no other token, including the action tokens, attends to the code queries, so the outputs of the code queries never influence action prediction. During training, the gradient of the code prediction loss in \cref{eq:code} reaches the rest of the network only through the tokens attended by the code queries. Since the action tokens attend to the same tokens, the code prediction loss shapes the tactile representations used for action prediction. Allowing the action tokens to attend to the code queries would let the code prediction loss bypass the shared tactile representations (\cref{sec:ablation}).

\paragraph{Training objectives.}
Let $h_s$ denote the final-layer hidden state of $u_s$. The code prediction head consists of four per-token classifiers, where the $s$-th classifier maps $h_s$ to a distribution $p_s(\cdot\mid h_s)$ over the 16 values of $c_t^{(s)}$. The head is supervised by the code $c_t$ from the tokenizer with the code prediction loss, defined as
\begin{equation}
  \mathcal{L}_{\text{code}}=-\frac{1}{4}\sum_{s=1}^{4}\log p_s(c_t^{(s)}\mid h_s).
  \label{eq:code}
\end{equation}
Predicting tactile innovation also requires estimating the current contact state, since tactile innovation is a deviation from the current contact state. We therefore attach a contact-state prediction head to the tactile encoder, which predicts contact properties such as contact area, shear, and indentation depth and yields the contact-state loss $\mathcal{L}_{\text{contact}}$. The final training objective is defined as
\begin{equation}
  \mathcal{L}=\mathcal{L}_{\text{flow}}+\lambda_{\text{contact}}\mathcal{L}_{\text{contact}}+\lambda_{\text{code}}\mathcal{L}_{\text{code}},
  \label{eq:total}
\end{equation}
where $\mathcal{L}_{\text{flow}}$ is the flow matching loss of the VLA, and the weights $\lambda_{\text{contact}}$ and $\lambda_{\text{code}}$ are set such that each auxiliary loss equals 15\% of $\mathcal{L}_{\text{flow}}$ at the beginning of training.

\begin{table*}[t]
  \centering\footnotesize\setlength{\tabcolsep}{4pt}\renewcommand{\arraystretch}{1.2}
  \caption{\textbf{Comparison of \vla and baselines on UniVTAC.} All policies are trained on the same 50 demonstrations per task. We report closed-loop success rates (\%), with rows sorted by average success rate in ascending order. Bold and underlined values denote the best and second-best results, respectively.}
  \label{tab:main}
  \begin{tabular}{lcccccccc}
    \toprule
    Method & Tactile & \ih~$\uparrow$ & \itb~$\uparrow$ & \lb~$\uparrow$ & \lc~$\uparrow$ & \pok~$\uparrow$ & \pb~$\uparrow$ & Avg.~$\uparrow$ \\
    \midrule
    UniVTAC-ACT~\cite{zhao2023act,univtac2026} & input & 46 & 55 & 49 & 27 & 41 & 25 & 40.5 \\
    $\pi_{0.5}$~\cite{intelligence2025pi05} & none & 88 & 70 & 26 & 25 & 36 & 40 & 47.5 \\
    FTP-$\pi_{0.5}$~\cite{yuan2026ftp1} & input & 77 & \textbf{99} & 45 & 33 & 19 & \textbf{88} & 60.2 \\
    FTP-1~\cite{yuan2026ftp1} & input & 63 & \textbf{99} & 61 & \textbf{94} & 40 & \textbf{88} & 74.2 \\
    \nzero~\cite{neoteai2026n0vtla} & input & \textbf{99} & 90 & 24 & 81 & \underline{77} & \underline{81} & 75.3 \\
    SimpleTouch~\cite{simpletouch2026} & input + predict & 94 & 93 & \textbf{82} & 65 & 54 & 73 & \underline{76.8} \\
    \midrule
    \vla (ours) & input + innovation target & \underline{95} & \underline{95} & \underline{78} & \underline{88} & \textbf{94} & 75 & \textbf{87.5} \\
    \bottomrule
  \end{tabular}
\end{table*}

\paragraph{Training and deployment.}
The policy is initialized from the pretrained weights of the VLA and trained end-to-end on the demonstrations with \cref{eq:total}, where the codes are precomputed by the tokenizer of \cref{sec:teacher}. Since neither the tokenizer nor the two auxiliary heads are required for action prediction, all three are removed at deployment, and the deployed model is the VLA with a 60-token tactile branch.

%% file: sec/4_experiments.tex
\section{Experiments}
\label{sec:exp}
We describe the experimental setup in \cref{sec:setup}, compare \vla with visuo-tactile baselines in simulation in \cref{sec:main}, and ablate each design in \cref{sec:ablation}. \Cref{sec:touchoff} examines whether each policy relies on tactile input, \cref{sec:attribution} verifies that a substantial portion of the tactile change is predictable from vision and executed actions, and \cref{sec:real} evaluates \vla in the real world.

\subsection{Experimental Setup}
\label{sec:setup}
\paragraph{Benchmark and data.}
We adopt six tasks from the UniVTAC benchmark~\cite{univtac2026}, namely \ih, \itb, \lb, \lc, \pok, and Put Bottle in Shelf (\pb), simulated in Isaac Sim with two vision-based tactile sensors~\cite{yuan2017gelsight} rendered by Taxim~\cite{si2022taxim}. Each task provides 100 expert demonstrations, of which demonstrations 0 to 49 are used to train the policies, 0 to 89 to train the tokenizer, and 90 to 99 are held out for analysis. We further set up four real-world tasks, namely wiping a board, inserting a plug into a power strip, lifting a lying can upright, and unscrewing a bottle cap, with a Xense vision-based tactile sensor on the fingertip of a Franka FR3 gripper (\cref{sec:real}). For each real-world task, we collect 50 demonstrations by teleoperation with GELLO~\cite{wu2024gello} at 30\,Hz, of which the first 45 train the policies and the tokenizer. Each policy is evaluated for 100 episodes per task with identical random seeds in simulation (\cref{tab:main,tab:ablation}) and for 50 trials per task in the real world.

\paragraph{Baselines.}
We compare \vla against five visuo-tactile baselines, namely UniVTAC-ACT, the visuo-tactile ACT provided by the benchmark~\cite{univtac2026,zhao2023act}, FTP-$\pi_{0.5}$ and FTP-1 (Foundation Tactile Policy)~\cite{yuan2026ftp1}, \nzero~\cite{neoteai2026n0vtla}, and SimpleTouch~\cite{simpletouch2026}. We also report $\pi_{0.5}$~\cite{intelligence2025pi05}, the VLA on which \vla is built, fine-tuned as a multi-task policy without tactile input. For fair comparison, all baselines are retrained on the same demonstrations with the official configurations, including the action execution horizon, and evaluated under the same protocol.

\subsection{Simulation Experiment}
\label{sec:main}
As shown in \cref{tab:main}, \vla achieves an average success rate of 87.5\% across the six tasks, 10.7 and 12.2 percentage points above the two strongest baselines, SimpleTouch and \nzero. Moreover, \vla is the only policy that reaches at least 75\% on every task, whereas each baseline drops to 54\% or below on at least one task. With the execution horizon of \vla, the baselines average 82.0\% (FTP-1), 79.3\% (SimpleTouch), 78.0\% (\nzero), and 68.0\% (FTP-$\pi_{0.5}$), all below the 91.3\% of \vla on the same seeds. Note that \vla requires no tactile pretraining, whereas FTP-1 and \nzero are initialized from policies pretrained on external visuo-tactile data, and SimpleTouch employs a frozen encoder pretrained on external tactile data.

\subsection{Ablation Study}
\label{sec:ablation}
\Cref{sec:intro} attributes the limited benefit of tactile input to the redundancy of the tactile signal and derives three designs from the attribution. To verify each design, \cref{tab:ablation} removes or alters one design of \vla at a time on the two lifting tasks, on which the tactile-off test distinguishes \vla from the baselines (\cref{sec:touchoff}). The variant without tactile input is the $\pi_{0.5}$ of \cref{tab:main}, and every other variant follows the training recipe, data, random seeds, and action execution horizon of \vla. All differences refer to the success rate averaged over the two tasks.

\paragraph{Impact of visual and action conditioning.}
To verify the necessity of conditioning, we remove the visual observations and executed actions from both the encoder and the decoder of the tokenizer, so that the code encodes the entire future tactile change (w/o conditioning), which lowers the average success rate by 21 points. Conditioning the tokenizer on the visual observations alone, as in a visual residual~\cite{zhang2026restacvla} (vision-only conditioning), recovers only part of the drop, and the average success rate remains 13.5 points lower. On held-out \lc demonstrations, the fraction of the tactile change that the decoder explains only through the code is 48\% without conditioning, 40\% with the visual observations alone, 23\% with the executed actions alone, and 13\% with both conditions. The action-induced change is therefore removed by the executed actions rather than by the visual observations, and continues to dominate a code conditioned on vision alone.

\paragraph{Impact of the multi-frame window.}
To examine whether the code must describe a sequence of tactile observations, we compute the code from only the two endpoints of the window, $t$ and $t{+}15$ (two-frame code), while keeping the policy unchanged. The average success rate drops by 6.5 points, since a code computed from the two endpoints misses the contact events that occur within the window.

\paragraph{Impact of the attention mask.}
To examine whether the code must serve as a training target rather than an input, we allow the action tokens to attend to the code queries (w/o attention mask). The average success rate drops by 15 points, demonstrating that the code is most effective when action prediction cannot access the code queries.

\paragraph{Impact of the tactile branch and the code prediction loss.}
Removing the tactile branch (w/o tactile input) and the code prediction loss (w/o code loss) reduces the average success rate by 57.5 and 7 points, respectively. The tactile branch provides the tactile signal, whereas the code prediction loss determines which part of the tactile signal the policy retains.

\begin{table}[t]
  \centering\footnotesize
  \caption{\textbf{Ablation study on the two lifting tasks.} We report success rates (\%). The variant w/o tactile input is the $\pi_{0.5}$ of \cref{tab:main}, and every other variant differs from \vla in one design.}
  \label{tab:ablation}
  \begin{tabular}{lccc}
    \toprule
    Variant & \lc~$\uparrow$ & \lb~$\uparrow$ & Avg.~$\uparrow$ \\
    \midrule
    \vla & \textbf{88} & \textbf{78} & \textbf{83.0} \\
    w/o conditioning & 63 & 61 & 62.0 \\
    vision-only conditioning & 62 & 77 & 69.5 \\
    two-frame code & 81 & 72 & 76.5 \\
    w/o attention mask & 66 & 70 & 68.0 \\
    w/o code loss & 84 & 68 & 76.0 \\
    w/o tactile input & 25 & 26 & 25.5 \\
    \bottomrule
  \end{tabular}
\end{table}

\begin{figure*}[t]
  \centering
  \includegraphics[width=0.9\textwidth]{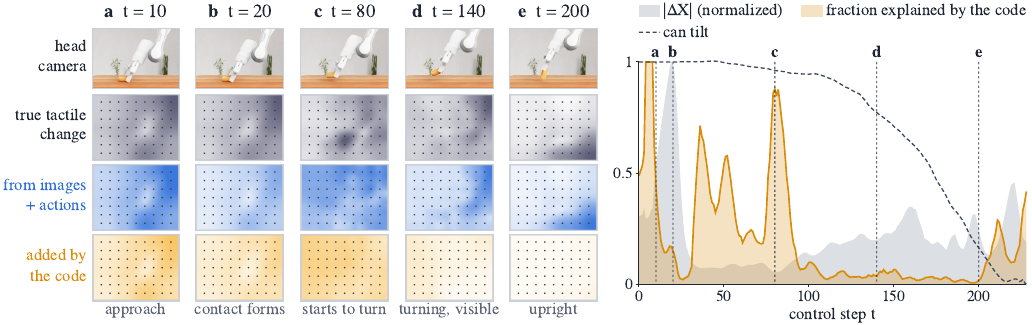}
  \caption{\textbf{Temporal distribution of tactile innovation on a held-out \lc demonstration.} The left panel shows, at five time steps, the head-camera image, the tactile change over the next 15 steps, and the components explained by visual observations and executed actions (blue) and by the code (orange). The right panel shows the magnitude of the tactile change (gray), the fraction explained by the code (orange), and the can tilt angle (dashed). The fraction explained by the code peaks before contact onset (a) and at incipient slip (c).}
  \label{fig:timeline}
\end{figure*}

\subsection{Analysis of Tactile Reliance}
\label{sec:touchoff}
A higher success rate does not by itself indicate that a policy relies on tactile input. We therefore apply the tactile-off test, which replaces the tactile input with a fixed no-contact frame at test time and keeps all other inputs unchanged, so that any change in the success rate is attributable solely to tactile input. \Cref{tab:touchoff} reports the results on the two lifting tasks, in which the object must be grasped firmly enough to be lifted yet loosely enough to rotate in the grasp. Removing tactile input reduces the success rate of \vla from 88\% to 0\% on \lc and to 4\% on \lb, whereas \nzero, FTP-1, and SimpleTouch retain most of their success rates, and FTP-1 even improves on \lb. Without tactile input, \vla reaches the can but fails to lift the can, whereas FTP-1 still completes the task. The open-loop results on \lc are consistent with the closed-loop results. On held-out demonstrations, the same replacement increases the errors of the arm and gripper actions predicted by \vla by 156\% and 238\%, respectively, far more than for any baseline. On tasks whose outcome depends on the grasp, the actions of \vla therefore depend on tactile input.

\begin{table}[t]
  \centering\footnotesize\setlength{\tabcolsep}{4pt}
  \caption{\textbf{Tactile-off test on the two lifting tasks.} The first two columns report the relative drop in closed-loop success rate, computed on the same seeds with and without tactile input (100 seeds on \lc and 25 on \lb, except 25 on both tasks for UniVTAC-ACT and 100 on both for \nzero). The last two columns report the relative change in the error of the predicted arm and gripper actions on held-out \lc demonstrations. Positive values indicate reliance on tactile input.}
  \label{tab:touchoff}
  \begin{tabular}{lcccc}
    \toprule
    & \multicolumn{2}{c}{\begin{tabular}{@{}c@{}}Closed-loop\\success drop\end{tabular}} & \multicolumn{2}{c}{\begin{tabular}{@{}c@{}}Open-loop action\\change, \lc\end{tabular}} \\
    \cmidrule(lr){2-3}\cmidrule(lr){4-5}
    Policy & \lc & \lb & Arm & Gripper \\
    \midrule
    \nzero~\cite{neoteai2026n0vtla} & 5\% & 8\% & $-$6\% & +4\% \\
    FTP-1~\cite{yuan2026ftp1} & 15\% & $-$27\% & $-$1\% & +45\% \\
    SimpleTouch~\cite{simpletouch2026} & 15\% & 5\% & +5\% & \underline{+84\%} \\
    UniVTAC-ACT~\cite{zhao2023act,univtac2026} & \underline{17\%} & \underline{58\%} & \underline{+9\%} & +50\% \\
    \vla (ours) & \textbf{100\%} & \textbf{95\%} & \textbf{+156\%} & \textbf{+238\%} \\
    \bottomrule
  \end{tabular}
\end{table}

\subsection{Analysis of Tactile Innovation}
\label{sec:attribution}
\paragraph{Predictability of the tactile change.}
To verify the decomposition in \cref{eq:innovation} at the dataset level, we train probes on the six tasks to predict the code without conditioning and the tactile innovation code, each from the context alone (task, visual observations, proprioceptive state, and executed future actions) or from the context together with the tactile history, and measure in nats, on the held-out demonstrations, the information that the tactile history adds beyond the context. For the code without conditioning, which encodes the entire future tactile change, the tactile history adds only 0.020 nats, so the context alone already accounts for 97\% of the information extracted with the tactile history. For the tactile innovation code, the tactile history adds 0.073 nats, 3.6$\times$ as much. The entire future tactile change, which tactile world models use as the prediction target, is therefore largely predictable without tactile input.

\paragraph{Temporal distribution of tactile innovation.}
Measured with the tokenizer on held-out demonstrations, the fraction of the tactile change that visual observations and executed actions cannot explain averages 37\% over the six tasks. On the held-out \lc demonstrations, the decoder output averaged over the distribution of codes explains 65\%, 78\%, and 87\% of the squared future tactile change at $+5$, $+10$, and $+15$ steps, respectively, and the code explains a further 25\%, 16\%, and 7\%. The fraction explained by the code peaks immediately before contact onset and at incipient slip, when the contact state is about to change, and vanishes once the motion of the can becomes visible in the visual observations (\cref{fig:timeline}).

\paragraph{Effect of the code prediction loss on the policy.}
We probe the tactile tokens in the action expert of \vla and of the variant without the code prediction loss, both fine-tuned on \lc, using per-token linear classifiers for $c_t$ on held-out \lc windows. The probe performs comparably on the tactile tokens of the two policies at the input layer of the action expert, but within the action expert the two policies differ on the 10\% of windows with the largest tactile innovation, where the tactile tokens of \vla predict the code with a top-1 accuracy of 49.0\%, compared with 38.6\% for the variant without the code prediction loss. Tactile input also modulates the attention of \vla. After accounting for the magnitude of the tactile change, the attention from the action tokens to the tactile tokens correlates with tactile innovation, and replacing the tactile input with a no-contact frame removes 90\% of the correlation, compared with 13\% for the variant without the code prediction loss. The code prediction loss therefore alters the tactile information that the action expert retains.

\subsection{Real-World Experiment}
\label{sec:real}
\begin{table}[t]
  \centering\footnotesize\setlength{\tabcolsep}{3pt}
  \caption{\textbf{Real-world experiments.} All policies are trained on the same demonstrations of each task, and we report success rates (\%). The visuo-tactile baselines are listed separately from the ablated settings and our method.}
  \label{tab:real}
  \begin{tabular}{lccccc}
    \toprule
    Policy & \begin{tabular}{@{}c@{}}Wipe\\Board~$\uparrow$\end{tabular} & \begin{tabular}{@{}c@{}}Insert\\Plug~$\uparrow$\end{tabular} & \begin{tabular}{@{}c@{}}Lift\\Can~$\uparrow$\end{tabular} & \begin{tabular}{@{}c@{}}Unscrew\\Cap~$\uparrow$\end{tabular} & Avg.~$\uparrow$ \\
    \midrule
    FTP-1~\cite{yuan2026ftp1} & \textbf{88} & 16 & 66 & 64 & 58.5 \\
    SimpleTouch~\cite{simpletouch2026} & 82 & 28 & 62 & 74 & 61.5 \\
    \midrule
    \vla, tactile-off & 52 & 10 & 46 & 52 & 40.0 \\
    w/o code loss & 62 & 22 & 50 & 56 & 47.5 \\
    w/o tactile input & 62 & 14 & 52 & 78 & 51.5 \\
    \midrule
    \vla (ours) & \textbf{88} & \textbf{38} & \textbf{74} & \textbf{82} & \textbf{70.5} \\
    \bottomrule
  \end{tabular}
\end{table}
We evaluate \vla on four contact-rich real-world tasks against two visuo-tactile baselines, FTP-1 and SimpleTouch, and three ablated settings of \vla, namely without tactile input, without the code prediction loss, and under the tactile-off test. All policies are trained on the same demonstrations. From the depth and marker estimates of the sensor, we compute the same indentation depth map and marker displacement field as in simulation. A trial is counted as successful when the writing is wiped off the board, the plug is fully inserted, the can stands upright, or the cap is separated from the bottle, with the same criteria for all policies. As shown in \cref{tab:real}, \vla achieves an average success rate of 70.5\%, surpassing SimpleTouch and FTP-1 by 9.0 and 12.0 percentage points, respectively, as well as all three ablated settings. The code prediction loss improves every real-world task by 16 to 26 percentage points. The comparable success rates of the variants without the code prediction loss (47.5\%) and without tactile input (51.5\%) indicate that, on the real-world tasks, the tactile branch is beneficial only when trained with tactile innovation as the target.

%% file: sec/5_conclusion.tex
\section{Conclusion}
In this work, we address the redundancy of the tactile signal in visuo-tactile VLA models, where a substantial portion of the tactile change is predictable from vision and executed actions. To this end, we introduce \method, which isolates tactile innovation and uses tactile innovation solely as a training target. Central to \method is a conditional multi-frame tokenizer that encodes tactile innovation into a 16-bit discrete code, which \vla predicts as an auxiliary objective. Experiments on six simulation tasks and four real-world tasks show that \vla achieves the highest average success rates among the evaluated visuo-tactile policies and relies on tactile input, and that removing the conditioning from the tokenizer lowers the success rate by 21 percentage points. The notion of innovation is not specific to tactile sensing, and extending \method to other modalities is left for future work.

%% file: sec/6_llm_statement.tex
\section*{LLM Usage Statement}
Claude was used to assist with coding, manuscript polishing, and result visualization. All AI-assisted outputs were reviewed and verified by the authors, who take full responsibility for this work.

%% file: sec/S_supp.tex
\section{Tasks and Data}
\label{sec:supp_data}
This section describes the simulation tasks, the demonstrations, and the data splits used in Sec.~4.

\begin{figure*}[t]
  \centering
  \includegraphics[width=\textwidth]{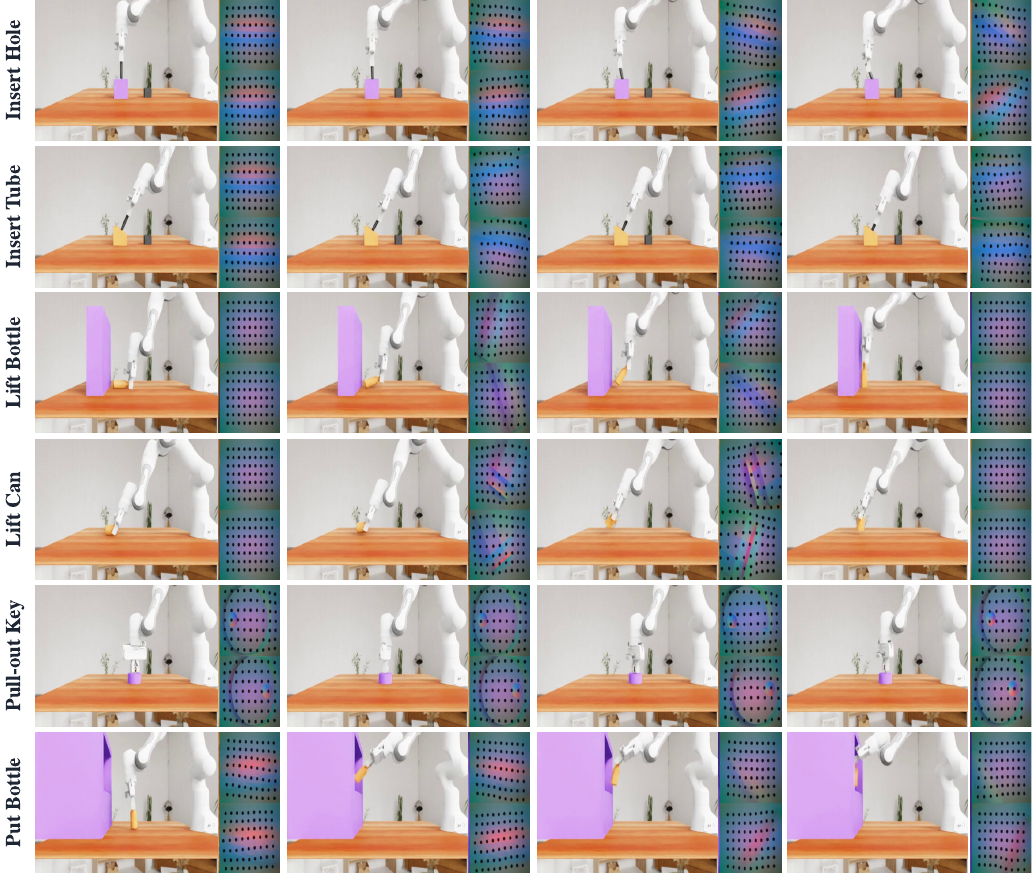}
  \caption{\textbf{Simulation rollouts of \vla.} Each row shows four frames of one successful rollout on one UniVTAC task from left to right, with the head-camera image and the images of the two tactile sensors of the same frame.}
  \label{fig:supp_sim_rollouts}
\end{figure*}

\paragraph{Simulation tasks.}
Following FTP-1~\cite{yuan2026ftp1} and SimpleTouch~\cite{simpletouch2026}, we evaluate on six tasks of the UniVTAC benchmark~\cite{univtac2026}, namely \ih, \itb, \lb, \lc, \pok, and \pb. The 100 demonstrations of each task are generated by the scripted expert of the benchmark in the simulator that is used for evaluation. \Cref{fig:supp_sim_rollouts} shows one successful rollout of \vla on each task.

\paragraph{Data splits.}
Demonstrations 0 to 49 of each task train every policy, including all baselines. Demonstrations 0 to 89 train the tokenizer, and demonstrations 90 to 99 are held out for the analyses in Secs.~4.3 to 4.5 and \cref{sec:supp_analysis}.

\paragraph{Real-world data.}
For each of the four real-world tasks, 50 demonstrations are collected by teleoperation, and the first 45 train the tokenizer and every policy (\cref{sec:supp_real}).

\section{Implementation Details}
\label{sec:supp_impl}
This section defines the tactile observation and the contact properties, and gives the architecture and the training details of the \method tokenizer and \vla. \Cref{tab:supp_hparam} lists the hyperparameters.

\subsection{Tactile Observation and Contact Properties}
\label{sec:supp_tactile}

\begin{figure*}[t]
  \centering
  \includegraphics[width=\textwidth]{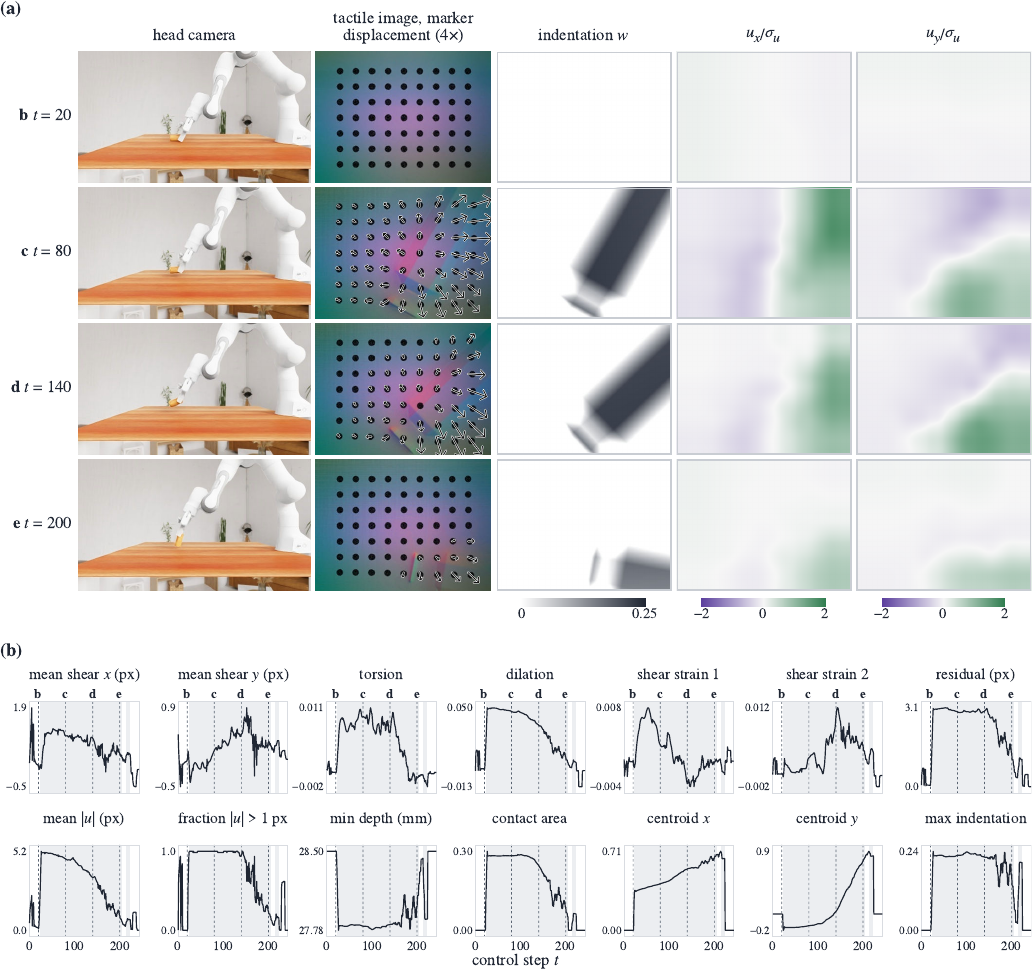}
  \caption{\textbf{Tactile observation and contact properties} on the held-out \lc demonstration of Fig.~4, left sensor. (a) At the steps b to e of Fig.~4, the head-camera image, the rendered tactile image with the marker displacements magnified four times, and the three channels of $X_t$ in \cref{eq:supp_obs}. (b) The 14 contact properties over the demonstration. Shaded intervals are in contact, and dashed lines mark the steps of (a).}
  \label{fig:supp_tactile}
\end{figure*}

\paragraph{Tactile observation.}
Each vision-based tactile sensor reports a depth map $D_t \in \mathbb{R}^{240\times320}$ of the gel surface in millimeters and the image positions of 63 markers arranged in a $7\times9$ grid. The observation of the right sensor is mirrored horizontally, so that both sensors share one image frame. The indentation depth map is
\begin{equation}
w_t(p) = \frac{1}{s}\,\min\bigl(\max\bigl(d_0 - D_t(p),\,0\bigr),\,w_{\max}\bigr),
\label{eq:supp_indent}
\end{equation}
where $p$ indexes the pixels, $d_0 = 28.5$\,mm, $w_{\max} = 6$\,mm, and $s = 3$\,mm. A pixel is in contact when $D_t(p) < 28.3$\,mm. The displacement of marker $m$ is $u_{t,m} = x_{t,m} - r_m$, where $x_{t,m}$ is its current image position and $r_m$ is its rest position, estimated as a median of the reference positions reported by the sensor. A marker is invalid when its reported reference position deviates from $r_m$ by 1 pixel or more, when its displacement is 80 pixels or more, or when it leaves the image, and the displacement of an invalid marker is filled by iterated averaging over its valid neighbors on the grid. The marker displacement field is
\begin{equation}
U_t = \frac{1}{\sigma_u}\,\mathcal{B}\bigl(u_{t,1},\dots,u_{t,63}\bigr) \in \mathbb{R}^{2\times96\times128},
\label{eq:supp_disp}
\end{equation}
where $\mathcal{B}$ interpolates the $7\times9$ grid of displacements bicubically onto a $96\times128$ grid, and $\sigma_u = 6.85$ pixels is the root mean square of the $x$ and $y$ components of the valid marker displacements on the contact frames of demonstrations 0 to 89. After $w_t$ is resampled to $96\times128$ by area averaging, the tactile observation stacks the three channels of the left sensor $l$ and the right sensor $r$,
\begin{equation}
X_t = \bigl[\,w_t^{(l)},\,U_t^{(l)}\,;\;w_t^{(r)},\,U_t^{(r)}\,\bigr] \in \mathbb{R}^{2\times3\times96\times128}.
\label{eq:supp_obs}
\end{equation}
\Cref{fig:supp_tactile}a shows the three channels of the left sensor at four steps of a held-out demonstration.

\paragraph{Contact properties.}
The contact-state loss of \vla (\cref{sec:supp_policy}) regresses 14 properties of each sensor. Nine properties summarize the marker displacements through the affine fit
\begin{equation}
(A_t, b_t) = \operatorname*{arg\,min}_{A,\,b} \sum_{m\in\mathcal{V}_t} \bigl\| u_{t,m} - A\,(r_m - \bar r) - b \bigr\|^2,
\label{eq:supp_affine}
\end{equation}
where $A \in \mathbb{R}^{2\times2}$, $b \in \mathbb{R}^{2}$, $\mathcal{V}_t$ is the set of valid markers, and $\bar r$ is the mean rest position. With $A = A_t$, the nine properties are the mean shear $b_t$ and
\begin{equation}
\begin{aligned}
\tau_t &= \tfrac{1}{2}\bigl(A_{21}-A_{12}\bigr), \qquad \delta_t = \tfrac{1}{2}\bigl(A_{11}+A_{22}\bigr),\\
\gamma^{(1)}_t &= \tfrac{1}{2}\bigl(A_{11}-A_{22}\bigr), \qquad \gamma^{(2)}_t = \tfrac{1}{2}\bigl(A_{12}+A_{21}\bigr),\\
\rho_t &= \Bigl(\tfrac{1}{|\mathcal{V}_t|}\textstyle\sum_{m}\bigl\|u_{t,m}-A_t(r_m-\bar r)-b_t\bigr\|^2\Bigr)^{1/2},\\
\bar u_t &= \tfrac{1}{|\mathcal{V}_t|}\textstyle\sum_{m}\|u_{t,m}\|,\\
\phi_t &= \tfrac{1}{|\mathcal{V}_t|}\bigl|\{m\in\mathcal{V}_t \mid \|u_{t,m}\| > 1\}\bigr|,
\end{aligned}
\label{eq:supp_props}
\end{equation}
namely the torsion $\tau_t$, the dilation $\delta_t$, the two shear strains $\gamma^{(1)}_t$ and $\gamma^{(2)}_t$, the non-affine residual $\rho_t$, the mean displacement magnitude $\bar u_t$, and the fraction $\phi_t$ of markers displaced by more than 1 pixel, where the sums run over $m \in \mathcal{V}_t$ and lengths are in pixels. Five properties are computed from the depth map, namely the minimum depth $\min_p D_t(p)$ clipped at $d_0$, the contact area as the fraction of pixels in contact, the two coordinates of the indentation-weighted centroid $\sum_p w_t(p)\,p\,/\sum_p w_t(p)$ in image coordinates normalized to $[-1, 1]$, and the maximum indentation $\max_p w_t(p)$. The contact label $e_t$ is 1 when any pixel is in contact and 0 otherwise. Each property is standardized with its mean and standard deviation on demonstrations 0 to 89 and clipped to $\pm30$, and frames in which fewer than half of the markers are valid do not contribute to the loss on the properties. \Cref{fig:supp_tactile}b plots the 14 properties over the demonstration.

\subsection{\method Tokenizer}
\label{sec:supp_tokenizer}

\paragraph{Inputs.}
The visual observation $o_t$ consists of the head-camera and wrist-camera images, and the tactile observation $X_t$ is given by \cref{eq:supp_obs}. The proprioceptive state $q_t \in \mathbb{R}^{8}$ contains the seven arm joint positions and the gripper opening. The executed action $a_{t+i}$ with $i = 1,\dots,15$ is the joint target executed at step $t{+}i$ minus $q_t$, standardized per step and per joint.

\paragraph{Tokens.}
The window consists of the frames $f = 0, 1, 2, 3$ at the steps $t{+}5f$. Each frame is represented by 137 tokens of width 384,
\begin{equation}
S_f = \bigl[\,H^{\text{tac}}_f;\; H^{\text{img}}_f;\; H^{\text{act}}_f;\; H^{\text{qry}}\,\bigr],
\label{eq:supp_frame_tokens}
\end{equation}
with $H^{\text{tac}}_f \in \mathbb{R}^{96\times384}$, $H^{\text{img}}_f \in \mathbb{R}^{32\times384}$, $H^{\text{act}}_f \in \mathbb{R}^{5\times384}$, and $H^{\text{qry}} \in \mathbb{R}^{4\times384}$. The tactile tokens $H^{\text{tac}}_f$ are produced by a $16\times16$ convolutional patch embedding of each sensor of $X_{t+5f}$, with position and sensor embeddings. The image tokens $H^{\text{img}}_f$ are the patch features of a frozen DINOv2-Base~\cite{oquab2024dinov2} for the two camera views of $o_{t+5f}$, each pooled to $4\times4$, projected by a LayerNorm and a linear layer, and added to a position embedding. The action tokens $H^{\text{act}}_f$ embed the five executed actions $a_{t+5f-4},\dots,a_{t+5f}$ that lead to frame $f$ with a shared two-layer MLP and a position embedding, and frame 0 uses five learnable padding tokens instead. $H^{\text{qry}}$ holds the four learnable queries.

\paragraph{Encoder.}
The encoder consists of 6 spatial-temporal blocks~\cite{bruce2024genie} with 6 attention heads. Each block applies spatial attention among the 137 tokens of a frame, causal temporal attention across frames at each token position, and an MLP. The temporal attention from frame $f$ to frame $f'$ uses the mask
\begin{equation}
M^{\text{temp}}_{f f'} = \begin{cases} 0, & f' \le f,\\ -\infty, & f' > f,\end{cases}
\label{eq:supp_causal}
\end{equation}
so the order of the frames is given by the mask and no temporal position embedding is used. The outputs of the four queries of frame 3 pass through a LayerNorm and a linear layer from 384 to 32 dimensions, which yields $z_t = (z_t^{(1)},\dots,z_t^{(4)}) \in \mathbb{R}^{4\times32}$. Each vector is quantized to the nearest entry of its codebook $\mathcal{C}_s = \{e^{(s)}_1,\dots,e^{(s)}_{16}\} \subset \mathbb{R}^{32}$,
\begin{equation}
\begin{aligned}
c_t^{(s)} &= \operatorname*{arg\,min}_{k\in\{1,\dots,16\}} \bigl\|z_t^{(s)} - e^{(s)}_k\bigr\|_2,\\
\hat z_t^{(s)} &= z_t^{(s)} + \operatorname{sg}\bigl(e^{(s)}_{c_t^{(s)}} - z_t^{(s)}\bigr),
\end{aligned}
\label{eq:supp_vq}
\end{equation}
where $\operatorname{sg}$ stops the gradient, so that the decoder receives the codebook entry and the encoder receives the gradient of $\hat z_t^{(s)}$ unchanged (straight-through estimator~\cite{oord2017vqvae}).

\paragraph{Decoder.}
The decoder has the same structure as the encoder. Frame 0 holds the tactile tokens of $X_t$, and each future frame holds 96 mask tokens in place of the tactile tokens, formed from one shared learnable vector with position and sensor embeddings. The four quantized vectors $\hat z_t^{(s)}$ are projected from 32 to 384 dimensions, added to token embeddings, and inserted into every frame. A LayerNorm and a linear output layer map each mask token of frame $k$ to a $16\times16\times3$ patch of the cumulative tactile change $\widehat{\Delta X}_{t:t+5k}$.

\paragraph{Objective.}
The tokenizer is trained with $\mathcal{L}_{\text{rec}} + \mathcal{L}_{\text{cb}} + 0.25\,\mathcal{L}_{\text{commit}}$, where the codebook and commitment losses of the VQ-VAE~\cite{oord2017vqvae} are
\begin{equation}
\begin{aligned}
\mathcal{L}_{\text{cb}} &= \tfrac{1}{4}\textstyle\sum_{s=1}^{4}\bigl\|\operatorname{sg}(z_t^{(s)}) - e^{(s)}_{c_t^{(s)}}\bigr\|^2,\\
\mathcal{L}_{\text{commit}} &= \tfrac{1}{4}\textstyle\sum_{s=1}^{4}\bigl\|z_t^{(s)} - \operatorname{sg}(e^{(s)}_{c_t^{(s)}})\bigr\|^2,
\end{aligned}
\label{eq:supp_vqloss}
\end{equation}
with the squared norms averaged over the 32 dimensions. The codebooks are initialized from encoder outputs, and unused entries are re-initialized every 2{,}000 steps. The reconstruction loss is the contact-weighted $L_1$ loss
\begin{equation}
\mathcal{L}_{\text{rec}} = \frac{1}{3}\sum_{k=1}^{3} \frac{\sum_{p} \omega_{k}(p)\,\bigl|\widehat{\Delta X}_{t:t+5k}(p) - \Delta X_{t:t+5k}(p)\bigr|}{\sum_{p} \omega_{k}(p)},
\label{eq:supp_rec}
\end{equation}
where $p$ indexes the pixels and channels of both sensors. The weight is $\omega_k(p) = 5$ when pixel $p$ is in contact at $t$ or at $t{+}5k$ and $\omega_k(p) = 1$ otherwise, where a pixel of the $96\times128$ grid is in contact when its indentation exceeds $(d_0 - 28.3\,\text{mm})/s$.

\paragraph{Training and labeling.}
For the simulation tasks, the tokenizer is trained on demonstrations 0 to 89 of the six tasks, and demonstrations 90 to 99 are held out. For the real-world tasks, the tokenizer is trained on the same 45 demonstrations per task as the policies. The trained tokenizer labels every demonstration frame whose step $t{+}15$ lies within the episode.

\begin{table}[t]
  \centering\footnotesize\setlength{\tabcolsep}{3pt}
  \caption{\textbf{Hyperparameters of the \method tokenizer and \vla.}}
  \label{tab:supp_hparam}
  \begin{tabular}{ll}
    \toprule
    Setting & Value \\
    \midrule
    \multicolumn{2}{l}{\textit{\method tokenizer}} \\
    Frames per window & 4 ($t$, $t{+}5$, $t{+}10$, $t{+}15$) \\
    Tactile, image tokens & 96, 32 per frame \\
    Action tokens, queries & 5, 4 per frame \\
    Token width, blocks, heads & 384, 6, 6 \\
    Code & 4 tokens, 16 entries each \\
    Code dimension & 32 \\
    Commitment weight & 0.25 \\
    Contact weight in $\mathcal{L}_{\text{rec}}$ & 5 \\
    Optimizer & AdamW, $\beta_2 = 0.95$, weight decay 0.05 \\
    Gradient clipping & 1.0 \\
    Learning rate, warm-up & $3\times10^{-4}$, 1{,}000 steps \\
    Learning rate decay & cosine, to $3\times10^{-5}$ \\
    Batch size, steps & 256, 12k \\
    \midrule
    \multicolumn{2}{l}{\textit{\vla}} \\
    Action chunk & 32 steps \\
    Proprioceptive, tactile tokens & 8, 60 \\
    Code queries & 4 \\
    $\lambda_{\text{contact}}$, $\lambda_{\text{code}}$ & 0.467, 0.119 \\
    Optimizer & AdamW, $\beta_2 = 0.95$ \\
    Gradient clipping & 1.0 \\
    Batch size & 64 \\
    Fine-tuning steps & 17k to 20k \\
    Peak learning rate & $5\times10^{-6}$ to $5\times10^{-5}$ \\
    Learning rate decay & cosine, to $5\times10^{-6}$ \\
    Warm-up & 200 to 500 steps \\
    Weight averaging & EMA, decay 0.99 \\
    \bottomrule
  \end{tabular}
\end{table}

\subsection{\vla}
\label{sec:supp_policy}

\paragraph{Base model.}
\vla is built on $\pi_{0.5}$~\cite{intelligence2025pi05}. The prefix is encoded by PaliGemma~\cite{beyer2024paligemma}, which combines a SigLIP~\cite{zhai2023siglip} vision encoder with a Gemma 2B~\cite{gemma2024} language model. The action expert is a Gemma 300M Transformer with 18 layers and a width of 1024. The policy predicts an action chunk of $H = 32$ steps, in which the arm joint targets are expressed relative to the current joint positions and the gripper target is the absolute opening.

\paragraph{Flow matching.}
As in $\pi_{0.5}$, the action expert is trained with flow matching~\cite{lipman2023flow} and predicts a velocity field $v_\theta$ that transports Gaussian noise to the action chunk $\mathbf{a}_t = a_{t+1:t+H}$. During training, a noisy chunk is formed as
\begin{equation}
\mathbf{a}^\tau_t = \tau\,\epsilon + (1-\tau)\,\mathbf{a}_t,\qquad \epsilon\sim\mathcal{N}(0, I),
\label{eq:supp_noisy}
\end{equation}
where $\tau = 0.999\,\beta + 0.001$ with $\beta \sim \mathrm{Beta}(1.5, 1)$, and the flow matching loss is
\begin{equation}
\mathcal{L}_{\text{flow}} = \mathbb{E}\,\bigl\| v_\theta(\mathbf{a}^\tau_t, \tau \mid \mathcal{O}_t) - (\epsilon - \mathbf{a}_t) \bigr\|^2,
\label{eq:supp_flow}
\end{equation}
averaged over the action dimensions of the robot, where $\mathcal{O}_t$ collects the prefix, the condition tokens, and the current state $q_t$, which modulates every layer of the action expert through adaptive RMS normalization together with the denoising time. At deployment, the chunk is obtained from $\mathbf{a}^{1} = \epsilon$ by ten Euler steps,
\begin{equation}
\mathbf{a}^{\tau-\Delta} = \mathbf{a}^{\tau} - \Delta\, v_\theta(\mathbf{a}^{\tau}, \tau \mid \mathcal{O}_t),\qquad \Delta = 0.1,
\label{eq:supp_euler}
\end{equation}
for $\tau = 1, 0.9, \dots, 0.1$, and $\mathbf{a}^{0}$ is the predicted chunk.

\paragraph{Condition tokens.}
Three groups of tokens are inserted before the noisy action tokens. The first group contains 8 proprioceptive tokens, which encode the joint positions from $t{-}7$ to $t$ with an MLP and a temporal embedding. The proprioceptive tokens serve only as keys and values and are not updated across layers. The second group contains 60 tactile tokens computed from the tactile observations at $t{-}8$, $t{-}4$, and $t$. For each frame and each sensor, the tactile encoder splits the observation into 48 patches of width 384, adds a summary query, and applies a two-layer pre-norm Transformer with 6 heads. The summary output and the patch outputs average-pooled to $3\times3$ give 10 tokens per sensor and frame. With $E_{\text{tac}}$ denoting this encoder followed by the pooling, the tactile tokens are
\begin{equation}
\begin{aligned}
H^{\text{tac}}_t = \tanh(g)\,\bigl(&W\,\mathrm{LN}\bigl(E_{\text{tac}}(X_{t-8}, X_{t-4}, X_t)\bigr)\\
&+ e^{\text{tok}} + e^{\text{frame}}\bigr) \in \mathbb{R}^{60\times1024},
\end{aligned}
\label{eq:supp_tac_tokens}
\end{equation}
where $W$ is a linear layer from 384 to 1024 dimensions, $e^{\text{tok}}$ and $e^{\text{frame}}$ are learnable token and frame embeddings, and the scalar gate $g$ is learnable and initialized to 0.1. Each group of condition tokens is scaled by its own gate of this form. The third group contains the four learnable code queries. The tactile tokens and the code queries are updated in every layer of the action expert.

\paragraph{Attention mask.}
Let $\mathcal{P}$ denote the prefix tokens, $\mathcal{H}$ the proprioceptive and tactile tokens, $\mathcal{U}$ the code queries, and $\mathcal{A}$ the noisy action tokens. In the action expert, a token $i$ attends to a token $j$ if and only if $j \in \mathcal{N}(i)$, where
\begin{equation}
\mathcal{N}(i) = \begin{cases}
\mathcal{P}\cup\mathcal{H}, & i\in\mathcal{H},\\
\mathcal{P}\cup\mathcal{H}\cup\mathcal{U}, & i\in\mathcal{U},\\
\mathcal{P}\cup\mathcal{H}\cup\mathcal{A}, & i\in\mathcal{A}.
\end{cases}
\label{eq:supp_mask}
\end{equation}
The code queries are thus attended only by the code queries themselves. In particular, the tactile tokens are prevented from attending to the code queries, because tactile tokens that are updated in every layer would otherwise relay the content of the code queries to the action tokens.

\paragraph{Code prediction head.}
The final-layer states of the four code queries pass through a LayerNorm and a per-token linear layer from 1024 to 16 dimensions. The code prediction loss is the cross-entropy with $c_t$, averaged over the four tokens and over the frames that carry a code.

\paragraph{Contact-state prediction head.}
For each frame and each sensor, the summary output of the tactile encoder, taken before the gate, passes through a linear layer from 384 to 256 dimensions, a GELU, and a linear layer from 256 to 15 dimensions. The first 14 outputs $\hat\mu_{f,g}$ regress the standardized contact properties $\mu_{f,g}$ of \cref{sec:supp_tactile} for frame $f \in \{t{-}8, t{-}4, t\}$ and sensor $g$, and the last output $\hat\kappa_{f,g}$ is a logit for the contact label $e_{f,g}$. All targets are computed from the sensor readings and require no additional annotation. The contact-state loss is
\begin{equation}
\mathcal{L}_{\text{contact}} = \frac{\sum_{f,g} \nu_{f,g}\,\tfrac{1}{14}\bigl\|\hat\mu_{f,g} - \mu_{f,g}\bigr\|^2}{\sum_{f,g}\nu_{f,g}} + \frac{1}{6}\sum_{f,g}\mathrm{BCE}\bigl(\hat\kappa_{f,g},\,e_{f,g}\bigr),
\label{eq:supp_contact}
\end{equation}
where $\nu_{f,g} \in \{0,1\}$ marks frames with at least half of the markers valid and BCE is the binary cross-entropy with logits. The gradient of $\mathcal{L}_{\text{contact}}$ reaches only the tactile encoder and the head.

\paragraph{Loss weights.}
We first train for 200 steps with both weights set to 1 and average each loss over steps 100 to 199, which gives $\bar{\mathcal{L}}_{\text{flow}} = 1.849$, $\bar{\mathcal{L}}_{\text{contact}} = 0.5935$, and $\bar{\mathcal{L}}_{\text{code}} = 2.338$. Each auxiliary term is then set to 15\% of $\mathcal{L}_{\text{flow}}$,
\begin{equation}
\lambda_{k} = 0.15\,\bar{\mathcal{L}}_{\text{flow}}\,/\,\bar{\mathcal{L}}_{k},\qquad k \in \{\text{contact}, \text{code}\},
\label{eq:supp_lambda}
\end{equation}
which yields $\lambda_{\text{contact}} = 0.467$ and $\lambda_{\text{code}} = 0.119$.

\paragraph{Training.}
The policy is fine-tuned on demonstrations 0 to 49 of each task with an exponential moving average of the weights (\cref{tab:supp_hparam}).

\paragraph{Deployment.}
The tokenizer, the code prediction head, and the contact-state prediction head are not executed at deployment. The four code queries remain in the sequence so that the positional encoding of the action tokens is unchanged, and no other token attends to the code queries. The action expert generates each action chunk with the ten steps of \cref{eq:supp_euler}. The condition tokens do not attend to the noisy action tokens and do not receive the denoising time, so the keys and values of the condition tokens are identical in all ten steps. These keys and values can therefore be computed once per inference and reused in every denoising step, which we do for the timing in \cref{tab:supp_cost}.

\subsection{Computational Cost}
\label{sec:supp_cost}
\Cref{tab:supp_cost} reports the cost of \vla on one NVIDIA H200 GPU with a batch size of 1. The deployed model has 5.38M more parameters than $\pi_{0.5}$, which belong to the tactile branch and the condition tokens. The code prediction head and the contact-state prediction head hold 0.17M parameters in total and are used only during training. Because both prediction heads are removed at deployment, the auxiliary objectives add no inference cost. Training the tokenizer takes 1.7 GPU hours, compared with about 10 GPU hours for one training run of the policy.

\begin{table}[t]
  \centering\footnotesize\setlength{\tabcolsep}{4pt}
  \caption{\textbf{Computational cost.} Inference time is the median over 50 forward passes of one action chunk, including observation preprocessing.}
  \label{tab:supp_cost}
  \begin{tabular}{lccc}
    \toprule
    Model & Deployed & Training-only & Inference \\
          & parameters & parameters & time (ms) \\
    \midrule
    $\pi_{0.5}$~\cite{intelligence2025pi05} & 3.479B & n/a & 173 \\
    \vla & 3.484B & 0.17M & 212 \\
    \bottomrule
  \end{tabular}
\end{table}

\section{Baselines and Evaluation Protocol}
\label{sec:supp_setup}
This section details how the baselines of Tab.~1 are trained and how all policies are evaluated in simulation.

\paragraph{Baselines.}
Each baseline is trained with the code and the default configuration released by its authors on demonstrations 0 to 49 of each task, and is evaluated on the same seeds as \vla.

\paragraph{Action execution horizon.}
\vla predicts 32 actions at a time and executes a prefix of each chunk before predicting again. The multi-task $\pi_{0.5}$ executes the first 10 actions before predicting again. The baselines keep the action execution of their authors. FTP-1 and FTP-$\pi_{0.5}$~\cite{yuan2026ftp1} predict at every control step and ensemble the predictions of the 20 most recent steps. SimpleTouch~\cite{simpletouch2026} executes 50 actions before predicting again, and 25 actions on the two insertion tasks. \nzero~\cite{neoteai2026n0vtla} executes 50 actions before predicting again. UniVTAC-ACT~\cite{zhao2023act,univtac2026} predicts at every control step with temporal aggregation.

\paragraph{Evaluation protocol.}
The evaluation seeds start from 1{,}000{,}000 and do not overlap with the seeds of the demonstrations. Episodes interrupted by simulator errors are discarded, so that every policy is evaluated on 100 valid episodes per task. Because the demonstrations are longer than those for which the default step limits of the benchmark were set, the step limit of every task is doubled for all policies.

\section{Additional Quantitative Results}
\label{sec:supp_results}
This section extends the tactile-off test of Sec.~4.4 to 100 seeds, reports the robustness of \vla to the training seed, and adds further ablations.

\subsection{Tactile-Off Test on 100 Seeds}
\label{sec:supp_touchoff}
\begin{figure*}[t]
  \centering
  \includegraphics[width=\textwidth]{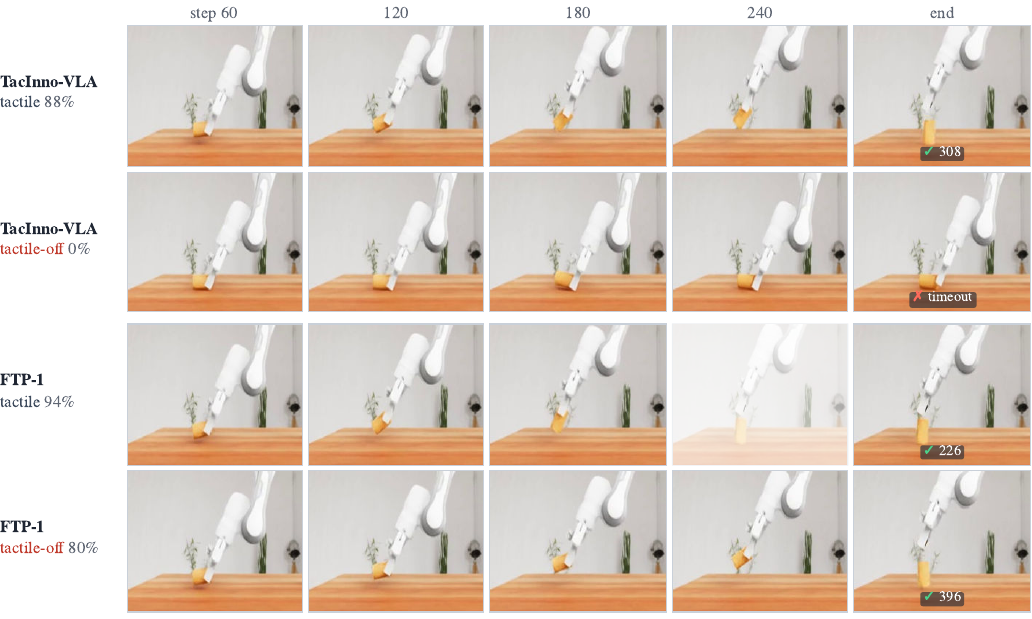}
  \caption{\textbf{Closed-loop tactile-off test on \lc.} Head-camera frames of the same seed with tactile input and under the tactile-off test. Without tactile input, \vla reaches the can but does not lift the can, whereas FTP-1 still completes the task. The percentage after each label is the success rate on 100 seeds, and the last column gives the step at which the task succeeds or the timeout.}
  \label{fig:supp_film}
\end{figure*}

\Cref{tab:supp_touchoff} repeats the closed-loop tactile-off test of Tab.~3 with 100 seeds for every policy on both lifting tasks. The conclusion is unchanged. Removing tactile input reduces the success rate of \vla from 88\% to 0\% on \lc and from 78\% to 8\% on \lb. \nzero loses at most 4 percentage points, FTP-1 retains 80\% on \lc and improves on \lb, and SimpleTouch loses 10 and 12 percentage points. \Cref{fig:supp_film} shows one seed of \lc, on which \vla reaches the can but does not lift the can without tactile input, whereas FTP-1 completes the task in both conditions.

\begin{table}[t]
  \centering\footnotesize\setlength{\tabcolsep}{4pt}
  \caption{\textbf{Tactile-off test with 100 seeds.} We report closed-loop success rates (\%) with tactile input and with the tactile input replaced by a fixed no-contact frame, on the same 100 seeds.}
  \label{tab:supp_touchoff}
  \begin{tabular}{lcccc}
    \toprule
    & \multicolumn{2}{c}{\lc} & \multicolumn{2}{c}{\lb} \\
    \cmidrule(lr){2-3}\cmidrule(lr){4-5}
    Policy & Tactile & Tactile-off & Tactile & Tactile-off \\
    \midrule
    \nzero~\cite{neoteai2026n0vtla} & 81 & 77 & 24 & 22 \\
    FTP-1~\cite{yuan2026ftp1} & 94 & 80 & 61 & 77 \\
    SimpleTouch~\cite{simpletouch2026} & 65 & 55 & 82 & 70 \\
    UniVTAC-ACT~\cite{zhao2023act,univtac2026} & 27 & 24 & 49 & 27 \\
    \vla (ours) & 88 & \textbf{0} & 78 & \textbf{8} \\
    \bottomrule
  \end{tabular}
\end{table}

\subsection{Robustness to the Training Seed}
\label{sec:supp_seeds}
We train \vla with three additional training seeds on every task and evaluate the resulting policies on up to 100 evaluation seeds per task, with the action execution horizon of Tab.~1. The six-task average success rates of the three additional seeds are 85.7\%, 91.1\%, and 86.7\%, compared with 87.5\% in Tab.~1. All three are above the 76.8\% of the strongest baseline in Tab.~1.

The drop of w/o attention mask in Tab.~2 is also robust to the training seed. We train this variant on \lc with two additional training seeds and evaluate it on the 100 seeds of Tab.~2. Across three training seeds, w/o attention mask reaches 66\%, 71\%, and 56\% with a mean of 64.3\%, whereas \vla reaches 88\%, 82\%, 85\%, and 88\% across four training seeds with a mean of 85.8\%. The gap between the two means, 21.5 points, matches the gap of 22 points on \lc in Tab.~2.

\subsection{Further Ablations}
\label{sec:supp_ablation}
The contact-state loss rests on the premise in Sec.~3.3 that predicting tactile innovation requires an estimate of the current contact state. To test the contribution of this loss, we train a variant without the contact-state loss (w/o contact-state loss) with the training recipe, data, random seed, and action execution horizon of \vla, and evaluate the variant on the 100 seeds of Tab.~2. \Cref{tab:supp_aux} compares the variant with w/o code loss of Tab.~2. Removing the contact-state loss lowers the average success rate by 16.5 points and the success rate on \lc from 88\% to 54\%. Removing either auxiliary loss lowers the average success rate, so \vla benefits from both auxiliary losses.

\begin{table}[t]
  \centering\footnotesize
  \caption{\textbf{Ablation of the two auxiliary losses on the two lifting tasks.} We report success rates (\%) on the 100 seeds of Tab.~2. The row w/o code loss is taken from Tab.~2.}
  \label{tab:supp_aux}
  \begin{tabular}{lccc}
    \toprule
    Variant & \lc~$\uparrow$ & \lb~$\uparrow$ & Avg.~$\uparrow$ \\
    \midrule
    \vla & \textbf{88} & 78 & \textbf{83.0} \\
    w/o code loss & 84 & 68 & 76.0 \\
    w/o contact-state loss & 54 & \textbf{79} & 66.5 \\
    \bottomrule
  \end{tabular}
\end{table}

Tab.~2 compares \vla with a tokenizer without conditioning and with a tokenizer conditioned on the visual observations alone. To complete this comparison, we condition the tokenizer on the executed actions alone (action-only conditioning) and train the policy with the training recipe, data, random seed, and action execution horizon of \vla. \Cref{tab:supp_cond_policy} shows that action-only conditioning reaches an average success rate of 68.0\%, close to the 69.5\% of vision-only conditioning and 15 points below \vla. Conditioning on either the visual observations or the executed actions alone therefore recovers on average only part of the drop of w/o conditioning, and the policy benefits most when the tokenizer is conditioned on both.

\begin{table}[t]
  \centering\footnotesize\setlength{\tabcolsep}{4pt}
  \caption{\textbf{Conditioning of the tokenizer on the two lifting tasks.} We report success rates (\%) on the 100 seeds of Tab.~2. All rows except action-only conditioning are taken from Tab.~2.}
  \label{tab:supp_cond_policy}
  \begin{tabular}{lccc}
    \toprule
    Conditioning & \lc~$\uparrow$ & \lb~$\uparrow$ & Avg.~$\uparrow$ \\
    \midrule
    w/o conditioning & 63 & 61 & 62.0 \\
    vision-only conditioning & 62 & 77 & 69.5 \\
    action-only conditioning & 72 & 64 & 68.0 \\
    \vla (vision and action) & \textbf{88} & \textbf{78} & \textbf{83.0} \\
    \bottomrule
  \end{tabular}
\end{table}

\section{Additional Analyses and Visualizations}
\label{sec:supp_analysis}
This section extends the analysis of conditioning in Sec.~4.3 to all six tasks, details the probes of Sec.~4.5, and provides further visualizations of tactile innovation.

\subsection{Effect of Conditioning on Every Task}
\label{sec:supp_cond}
\begin{figure*}[t]
  \centering
  \includegraphics[width=\textwidth]{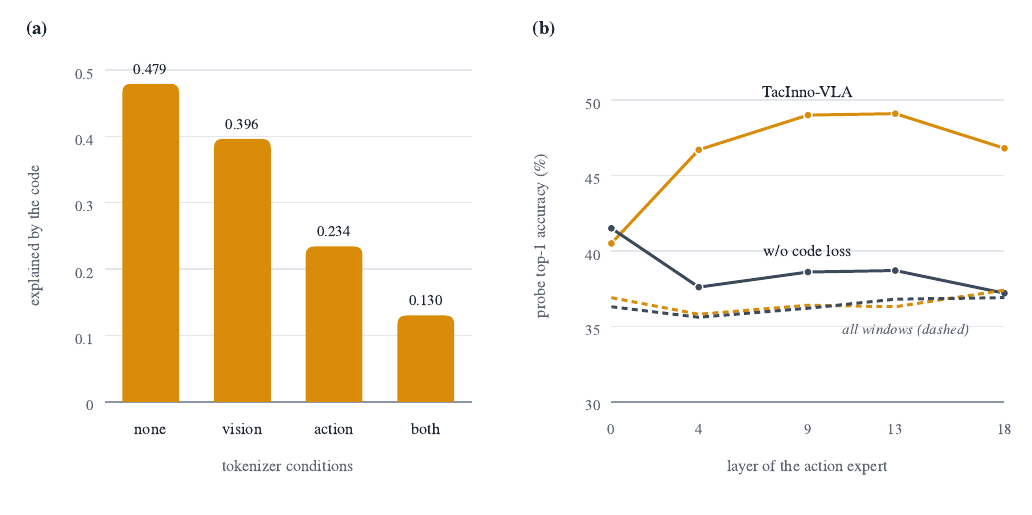}
  \caption{\textbf{Additional analyses on \lc.} (a) Fraction of the tactile change that the decoder explains only through the code, for four choices of the tokenizer conditions. (b) Top-1 accuracy of linear probes that predict the code from the tactile tokens at five depths of the action expert, on the 10\% of held-out windows with the largest tactile innovation (solid) and on all held-out windows (dashed).}
  \label{fig:supp_analysis}
\end{figure*}

Sec.~4.3 reports the effect of conditioning on \lc, and \cref{tab:supp_cond} extends the measurement to all six tasks. We train four tokenizers that differ only in the conditions. For a held-out window, the decoder output with the true code and the decoder output with the code averaged out are
\begin{equation}
\begin{aligned}
P_t &= \mathcal{D}\bigl(X_t, c_t;\, o_{t:t+15}, a_{t+1:t+15}\bigr),\\
M_t &= \frac{1}{K}\sum_{j=1}^{K} \mathcal{D}\bigl(X_t, \tilde c_j;\, o_{t:t+15}, a_{t+1:t+15}\bigr),
\end{aligned}
\label{eq:supp_PM}
\end{equation}
where the $K = 64$ codes $\tilde c_j$ draw each token $\tilde c_j^{(s)}$ independently from the frequency $p_s$ of its values on demonstrations 0 to 89 of the task. With all squared norms summed over the held-out windows, the three horizons, both sensors, the channels, and the pixels, the fraction of the future tactile change explained without the code and the fraction explained only through the code are
\begin{equation}
\begin{aligned}
\eta_M &= 1 - \frac{\sum\|\Delta X - M\|^2}{\sum\|\Delta X\|^2},\\
\eta_P - \eta_M &= \frac{\sum\|\Delta X - M\|^2 - \sum\|\Delta X - P\|^2}{\sum\|\Delta X\|^2},
\end{aligned}
\label{eq:supp_eta}
\end{equation}
where $\eta_P$ is defined as $\eta_M$ with $P$ in place of $M$. \Cref{tab:supp_cond} reports $\eta_M$. With visual and action conditioning, the decoder explains a larger fraction on every task than without conditioning, so the code needs to encode a smaller part of the tactile change. \Cref{fig:supp_analysis}a shows the fraction $\eta_P - \eta_M$ that is explained only through the code on \lc.

\begin{table}[t]
  \centering\footnotesize\setlength{\tabcolsep}{5pt}
  \caption{\textbf{Fraction of the future tactile change explained without the code (\%).} The fraction is measured on the held-out demonstrations with the code averaged out.}
  \label{tab:supp_cond}
  \begin{tabular}{lcc}
    \toprule
    Task & w/o conditioning & Visual and action \\
         &                  & conditioning \\
    \midrule
    \ih  & 20.6 & 55.0 \\
    \itb & 45.6 & 75.0 \\
    \lb  & 41.7 & 58.0 \\
    \lc  & 43.3 & 80.4 \\
    \pok & 16.5 & 41.1 \\
    \pb  & 40.9 & 74.1 \\
    \bottomrule
  \end{tabular}
\end{table}

\subsection{Predictability Probes}
\label{sec:supp_probe}
\paragraph{Probes of Sec.~4.5.}
Each probe is a small 4-layer predictor of width 384 that outputs a distribution over the 16 values of each of the four code tokens. The context-only probe receives a task token, the DINOv2 features of the camera images at $t{-}8$, $t{-}4$, and $t$, the joint positions from $t{-}7$ to $t$, and the executed actions from $t{+}1$ to $t{+}15$. The second probe additionally receives the tactile observations at $t{-}8$, $t{-}4$, and $t$. The probes are trained on demonstrations 0 to 49 of the six tasks, demonstrations 50 to 69 are used to select the training step and to calibrate the temperature, and demonstrations 90 to 99 are used for testing. With $q^{\text{ctx}}_s$ and $q^{\text{tac}}_s$ denoting the calibrated distributions of the two probes for token $s$, the information that the tactile history adds is the difference between their test cross-entropies, averaged over the six tasks,
\begin{equation}
\Delta I = \frac{1}{6}\sum_{k=1}^{6}\frac{1}{4}\sum_{s=1}^{4}\frac{1}{|\mathcal{T}_k|}\sum_{t\in\mathcal{T}_k}\log\frac{q^{\text{tac}}_s\bigl(c^{(s)}_t\bigr)}{q^{\text{ctx}}_s\bigl(c^{(s)}_t\bigr)},
\label{eq:supp_info}
\end{equation}
in nats per code token, where $\mathcal{T}_k$ is the set of test windows of task $k$.

\paragraph{Layer-wise probes of the policy.}
\Cref{fig:supp_analysis}b extends the probe of the tactile tokens in Sec.~4.5 to five depths of the action expert. At the input layer, the tactile tokens of \vla and of the variant without the code prediction loss predict the code comparably well. Within the action expert, the tactile tokens of \vla become more predictive of the code on the windows with the largest tactile innovation, whereas the tactile tokens of the variant do not. On all held-out windows, the two policies remain close. The difference between the two policies is therefore concentrated on the windows where the tactile change is least explained by vision and executed actions.

\subsection{Codebook Usage}
\label{sec:supp_codebook}
On the held-out demonstrations, the perplexities of the four code tokens are 11.2, 13.8, 10.9, and 10.3 out of 16, so every token uses most of the entries of its codebook.

\subsection{Further Visualizations}
\label{sec:supp_vis}
\begin{figure*}[t]
  \centering
  \includegraphics[width=\textwidth]{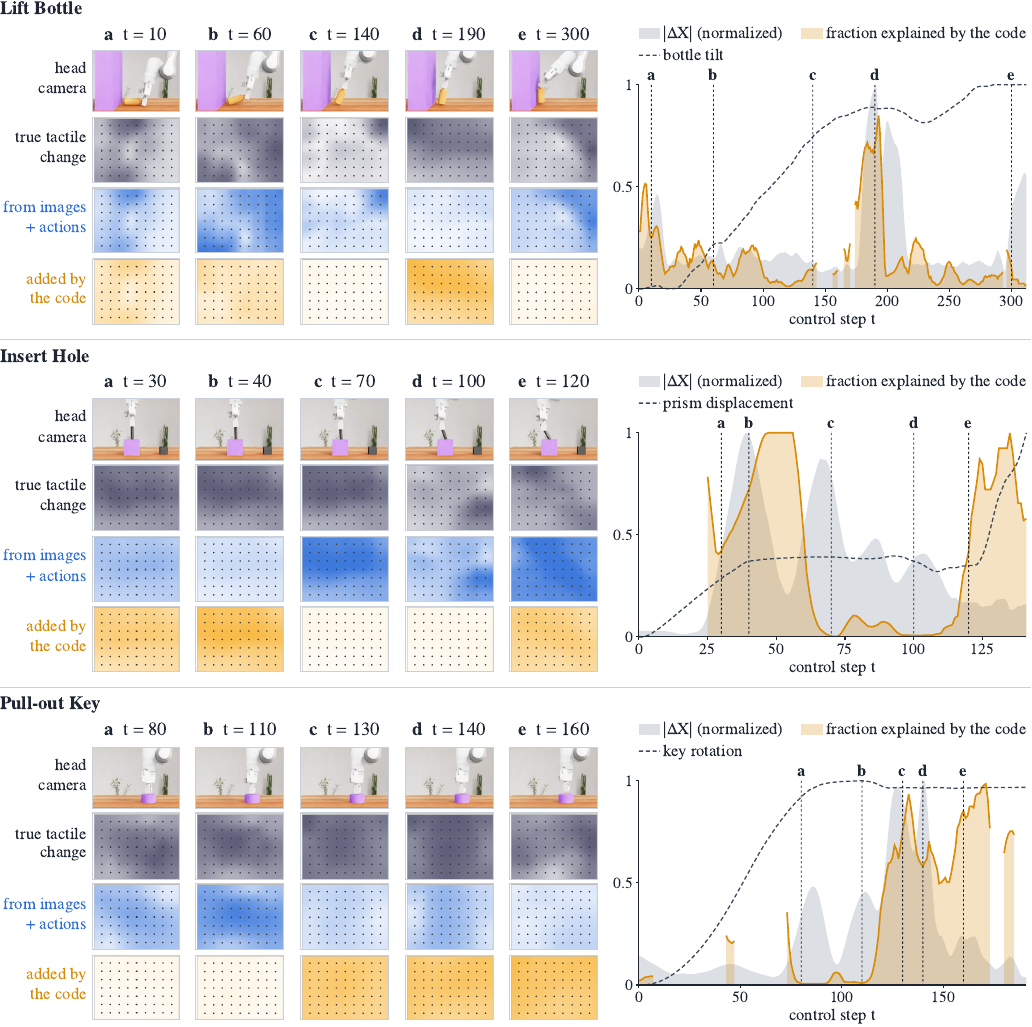}
  \caption{\textbf{Temporal distribution of tactile innovation on three further tasks.} Layout as in Fig.~4, for one held-out demonstration each of \lb, \ih, and \pok. The dashed line shows the tilt of the bottle, the displacement of the prism, and the rotation of the key, each normalized. The fraction explained by the code is omitted where the tactile change is below 10\% of its maximum in the demonstration.}
  \label{fig:supp_timeline}
\end{figure*}

\Cref{fig:supp_timeline} applies the analysis of Fig.~4 to one held-out demonstration each of \lb, \ih, and \pok. The fraction explained by the code at step $t$ is $\|P_t - M_t\|^2 / \|\Delta X_{t}\|^2$ with $P_t$ and $M_t$ from \cref{eq:supp_PM}, clipped to $[0, 1]$ and averaged over five steps, where the norms are summed over the three horizons, both sensors, the channels, and the pixels. The code adds little at c in \lb, at c and d in \ih, and at b in \pok, where visual observations and executed actions already explain most of the tactile change, and adds the most at d in \lb, at b in \ih, and from c to e in \pok. The three tasks thus show a pattern similar to that of \lc in Fig.~4.

\begin{figure*}[p]
  \centering
  \includegraphics[width=\textwidth]{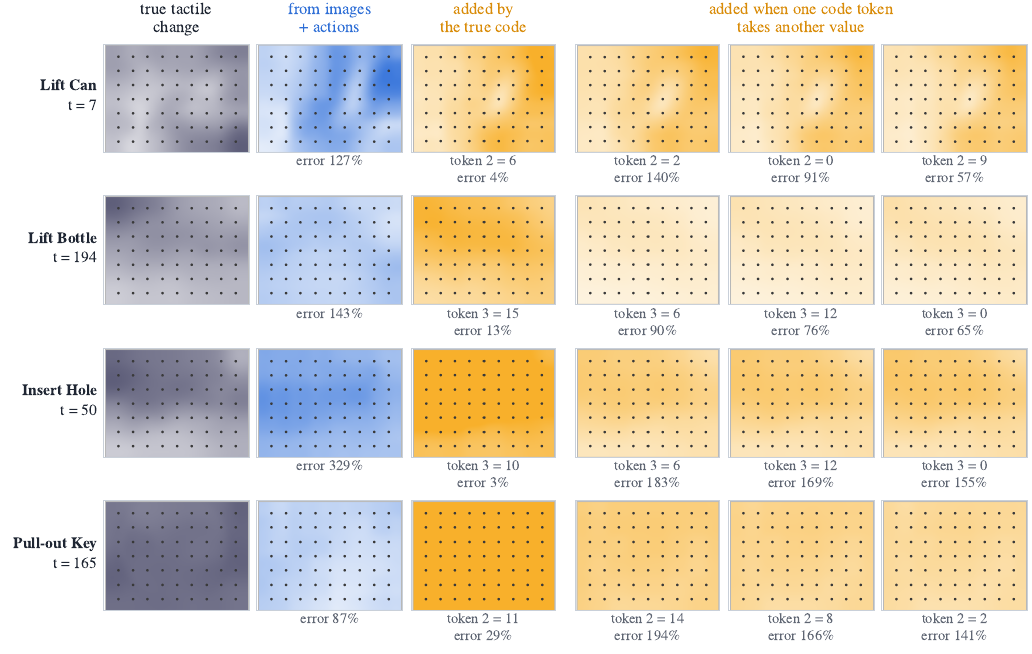}
  \caption{\textbf{What the code encodes.} In four held-out windows with fixed visual observations, executed actions, and tactile history, the columns show the tactile change over the next 15 steps, the component explained by vision and actions (blue), and the components added by the true code and by one changed code token (orange). The numbers give the squared error relative to the energy of the true change.}
  \label{fig:supp_code_swap}
\end{figure*}

\Cref{fig:supp_code_swap} illustrates what the code encodes. For four held-out windows in which visual observations and executed actions explain the tactile change poorly, we keep the visual observations, the executed actions, and the tactile history fixed and decode the tactile change with the true code and with one code token set to other values that occur on demonstrations 0 to 89. For a decoder output $\widehat{\Delta X}$ we report the relative error $\|\widehat{\Delta X} - \Delta X\|^2 / \|\Delta X\|^2$ on the displayed sensor. Visual observations and executed actions alone, through $M_t$, leave an error of 87\% to 329\% of the energy of the true change, and the true code reduces the error to 3\% to 29\%. Setting a single token to one of the values shown raises the error to between 57\% and 194\%. These examples show that the code carries information specific to the upcoming tactile change.

\begin{figure*}[p]
  \centering
  \includegraphics[width=\textwidth]{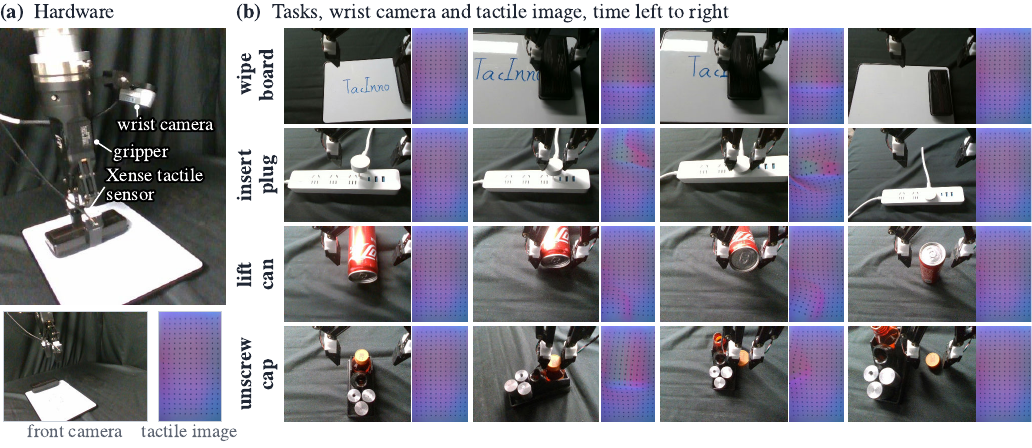}
  \caption{\textbf{Real-world setup and tasks.} (a) The Franka FR3 arm with a parallel gripper, a vision-based tactile sensor on the fingertip, a third-person camera, and a wrist camera. (b) Wrist-camera key frames and tactile images of one demonstration per task.}
  \label{fig:supp_real}
\end{figure*}

\begin{figure*}[t]
  \centering
  \includegraphics[width=\textwidth]{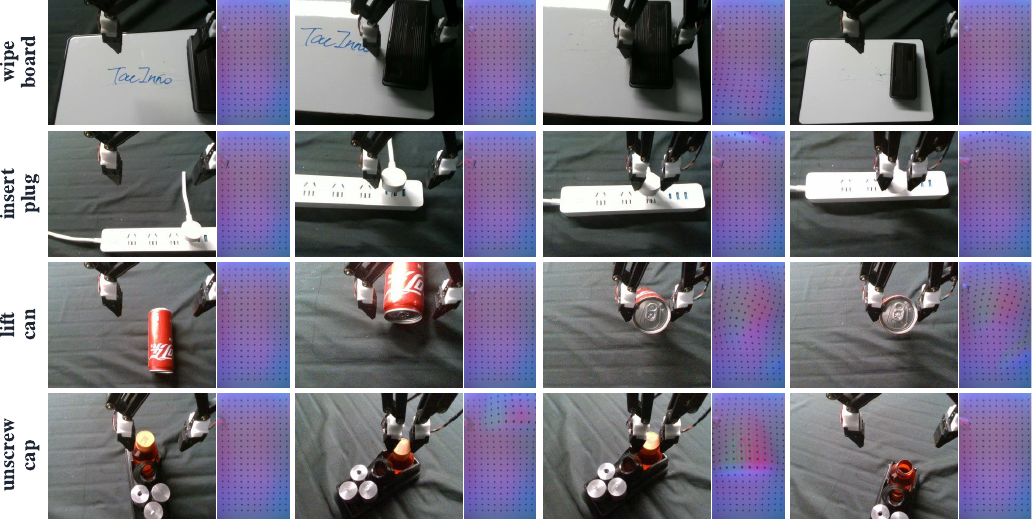}
  \caption{\textbf{Real-world rollouts of \vla.} Each row shows four steps of one successful trial from left to right, with the wrist-camera image and the tactile image of the same step.}
  \label{fig:supp_rollouts}
\end{figure*}

\section{Real-World Experiment Details}
\label{sec:supp_real}
This section describes the hardware, the tasks, and the data of the real-world experiment in Sec.~4.6.

\paragraph{Hardware.}
The real-world platform is a Franka FR3 arm with a parallel gripper, a Xense vision-based tactile sensor on the fingertip, a third-person RealSense camera, and a wrist-mounted RealSense camera (\cref{fig:supp_real}a).

\paragraph{Tasks and success criteria.}
The four tasks are wiping a board, inserting a plug into a power strip, lifting a lying can upright, and unscrewing a bottle cap (\cref{fig:supp_real}b). A trial is counted as successful when the writing is wiped off the board, the plug is fully inserted, the can stands upright, or the cap is separated from the bottle. The criteria are identical for all policies.

\paragraph{Data and training.}
For each task, 50 demonstrations are collected by teleoperation with GELLO~\cite{wu2024gello} at 30\,Hz, and the first 45 are used to train the tokenizer and every policy. The indentation depth map and the marker displacement field are computed from the depth and marker estimates of the sensor, in the same format as in simulation.

\paragraph{Trial protocol.}
Each policy is evaluated for 50 trials per task under the same protocol. When wiping the board and lifting the can, a trial ends at twice the median length of the demonstrations. A safety stop counts as a failure, and trials interrupted by hardware faults unrelated to the policy are repeated.

\paragraph{Rollouts.}
\Cref{fig:supp_rollouts} shows one successful trial of \vla on each real-world task.